\documentclass[10pt,twocolumn,letterpaper]{article}

\usepackage{iccv}
\usepackage{times}
\usepackage{epsfig}
\usepackage{graphicx}
\usepackage{amsmath}
\usepackage{amssymb}
\usepackage{makecell}
\usepackage{microtype}
\usepackage{footnote}
\usepackage{algorithm}
\usepackage{algpseudocode}

\usepackage{siunitx} 
\usepackage[pagebackref=true,breaklinks=true,letterpaper=true,colorlinks,bookmarks=false]{hyperref}
\usepackage{hypcap}

\newcommand{\vars}{\texttt}
\newcommand{\func}{\textrm}

\iccvfinalcopy 

\def\iccvPaperID{1218} 

\ificcvfinal\fi

\begin{document}

\title{Out-of-Core Surface Reconstruction via Global $TGV$ Minimization}

\author{Nikolai Poliarnyi\\
Agisoft LLC, St. Petersburg, Russia\\
{\tt\small polarnick@agisoft.com}
}

\maketitle
\ificcvfinal\thispagestyle{empty}\fi

\begin{abstract}
We present an out-of-core variational approach for surface reconstruction from a set of aligned depth maps.
Input depth maps are supposed to be reconstructed from regular photos or/and can be a representation of terrestrial LIDAR point clouds.
Our approach is based on surface reconstruction via total generalized variation minimization ($TGV$) because of its strong visibility-based noise-filtering properties and GPU-friendliness.
Our main contribution is an out-of-core OpenCL-accelerated adaptation of this numerical algorithm which can handle arbitrarily large real-world scenes with scale diversity.

\end{abstract}

\section{Introduction}

The structure from motion pipeline makes it possible to take photos of the same object/scene
and then not only align and calibrate these photos, but also reconstruct an observed surface with a high amount of details.
At the moment, the progress in camera sensor development opens a possibility for a regular user to take photos with a size up to hundreds of megapixels, the number which has been increasing rapidly over the past decades. Additionally, due to help of UAVs, affordable quadrocopters and automatic flight planners,
it becomes possible to gradually increase the amount of pictures one can take in a short span of time.
Therefore, in the area of photogrammetry, the task of being able to use all of the available data for a detailed noise-free surface reconstruction in an out-of-core fashion is necessary to make a highly detailed large scale reconstruction possible on affordable computers with limited RAM.

\begin{figure}
    \centering
    \capstart
    \begin{minipage}[b]{\linewidth}
        \includegraphics[width=\textwidth]{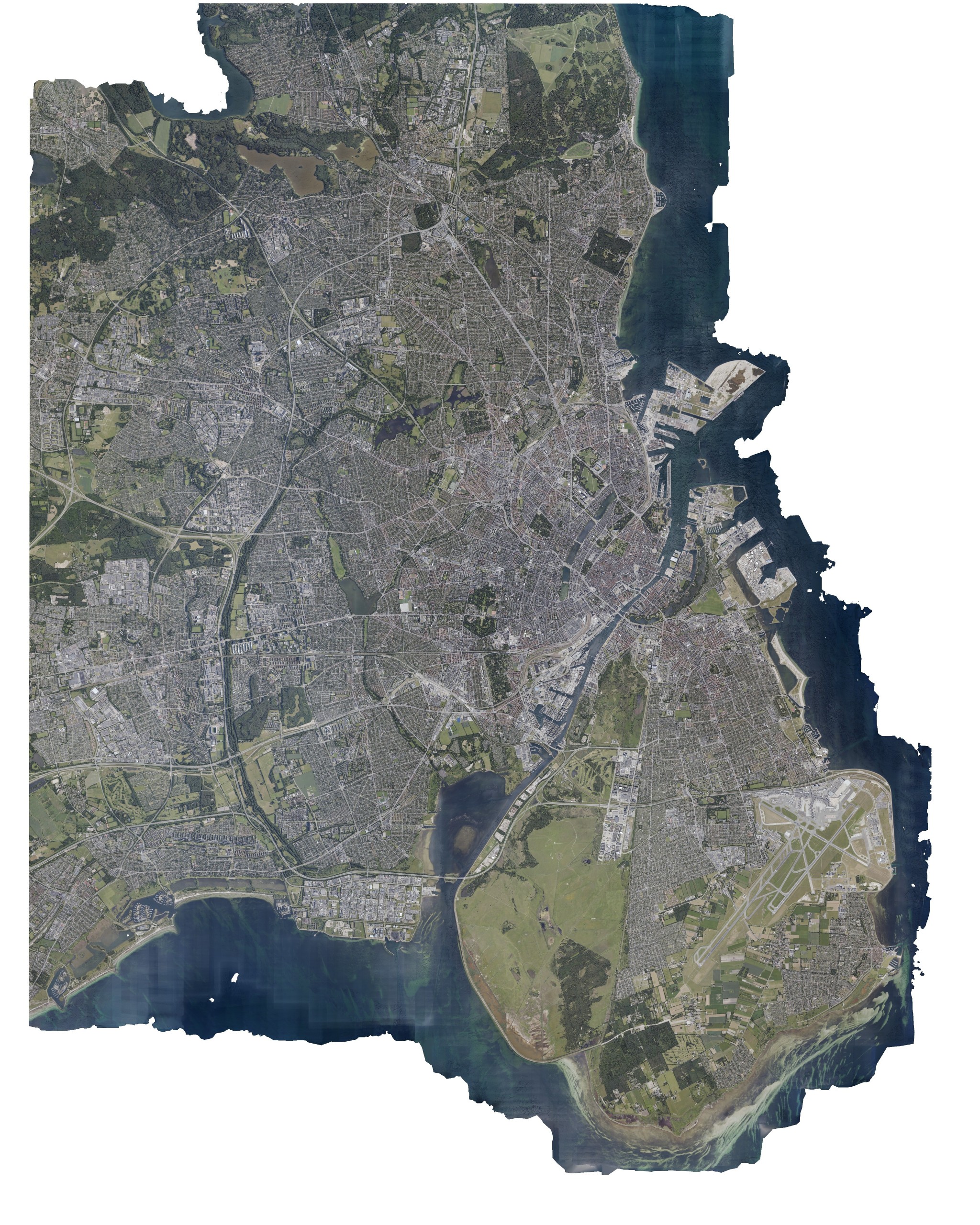}
    \end{minipage}
    \caption{Our method can handle an arbitrary large scene -- even \SI{425}{\km\squared} of the Copenhagen city
    (this polygonal model was reconstructed from 27472 aerial photos -- see supplementary for details).}
    \label{fig:copenhagen_full_rgb}
\end{figure}

We present a surface reconstruction method that has strong noise-filtering properties and can take both depth maps and terrestrial LIDAR scans as an input.
The whole method is implemented in an out-of-core way: the required memory usage is low even for very large datasets - we targeted the usage to be around 16 GB even for a Copenhagen city dataset with 27472 photos - see Fig.~\ref{fig:copenhagen_full_rgb}.
Each processing stage is divided into independent parts for out-of-core guarantees,
thus additionally obtaining a massive parallelism property (i.e. pipeline is cluster-friendly).
Calculation-heavy stages (the computation of histograms and iterative numeric scheme) are accelerated with GPUs via OpenCL API.


\section{Related Work}



\textit{Poisson surface reconstruction} method \cite{kazhdan2006poisson} performs well in local geometry details preservation by respecting normals of input point clouds, however it normally fails to handle scale diversity and often fails to filter noise between the surface and the sensor origins.
It should be noted that handling of scale diversity can be added, and noise filtering can be implemented on an early stage as a depth map filtering \cite{merrell2007real}. Nevertheless,
while Poisson reconstruction can be implemented in an out-of-core fashion \cite{bolitho2007multilevel} -
the preceding depth map filtering approach will likely require a high amount of memory to keep all depth maps relevant for a current depth map filtering in RAM.

\textit{Graph cut}-based reconstruction methods \cite{hiep2009towards}, \cite{jancosek2014exploiting}, \cite{han2019scalable}, \cite{zhou2019detail}
explicitly take into account visibility rays from sensors' origins to samples in depth maps, and therefore such methods have great noise filtering properties.
Scale diversity is also naturally supported via Delaunay tetrahedralized space discretization.
However, Delaunay tetrahedralization and minimum graph cut estimation of an irregular graph \cite{boykov2004experimental}, \cite{goldberg2015faster}
have high memory consumption and are computationally heavy. Because of this,
the out-of-core SSR \cite{mostegel2017scalable} method happens to be more than one order of magnitude slower than our method.

\textit{Local fusion} methods \cite{curless1996volumetric}, \cite{kuhn2015incremental}, \cite{kuhn2017tv} including FSSR \cite{fuhrmann2014floating}, \cite{fuhrmann2014mve}
are well suited for parallelization and scalability \cite{kuhn2015incremental}.
However on the other hand due to their local nature, they have weak hole filling properties and can not filter strong depth map noise
in difficult cases like the basin of the fountain in the Citywall dataset as shown in \cite{ummenhofer2015global}.

\textit{Photoconsistent mesh refinement}-based methods \cite{vu2011high}, \cite{li2016efficient} are fast due to GPU acceleration but are not able to change the topology of an input mesh,
and thus they heavily depend on the quality of an initial model reconstruction.

\textit{Total variation minimization}-based methods \cite{zach2007globally}, \cite{graber2011online}, \cite{pock2011tgv} are shown to have great noise filtering properties due to visibility constraints,
and can easily be GPU-accelerated \cite{zach2008fast}. Additionally, as shown in \cite{ummenhofer2015global}, \cite{ummenhofer2017global}, a variation minimization scheme can be implemented
in a compact and scale diversity-aware way by the use of a balanced octree, but even with such compact space representation, the peak memory consumption becomes critical for a large scale scene reconstruction task.

Our $TGV$-functional formulation follows \cite{pock2011tgv}, it was adapted for 3D space in a way, discussed in \cite{zach2007globally}.
We use a 2:1 balanced octree similar to \cite{ummenhofer2015global} for 3D space representation.
In contrast with their method, our framework has strict peak memory guarantees and is much faster thanks to GPU acceleration in the most time-consuming stages (as shown in \cite{ummenhofer2015global} -- its bottlenecks were in the computation of the histograms (17\%) and the energy minimization (80\%) stages, so we implemented both of them on GPU).

\section{Algorithm Overview}

To begin with, we would like to discuss the chosen functional minimization with a focus on noise robustness, scale diversity awareness and GPU-friendliness, while not taking memory requirements into account. Later, we will show how to adapt this minimization scheme for an out-of-core fashion in Section \ref{sec:out-of-core}.

\subsection{Distance Fields as Input Data}
\label{sec:dist-fields-as-input-data}

We prefer to be able to support different kinds of range image data as an input, such as:

\begin{itemize}
    \item Depth maps built with stereo methods like SGM \cite{hirschmuller2007stereo} from regular terrestrial photos or aerial UAV photos;
    \item RGB-D cameras, which are essentially the same as previously mentioned depth maps;
    \item Terrestrial LIDAR point clouds with a known sensor origin.
\end{itemize}

\begin{figure}
    \centering
    \begin{minipage}[b]{0.54\linewidth}
    \capstart
    \includegraphics[width=\textwidth]{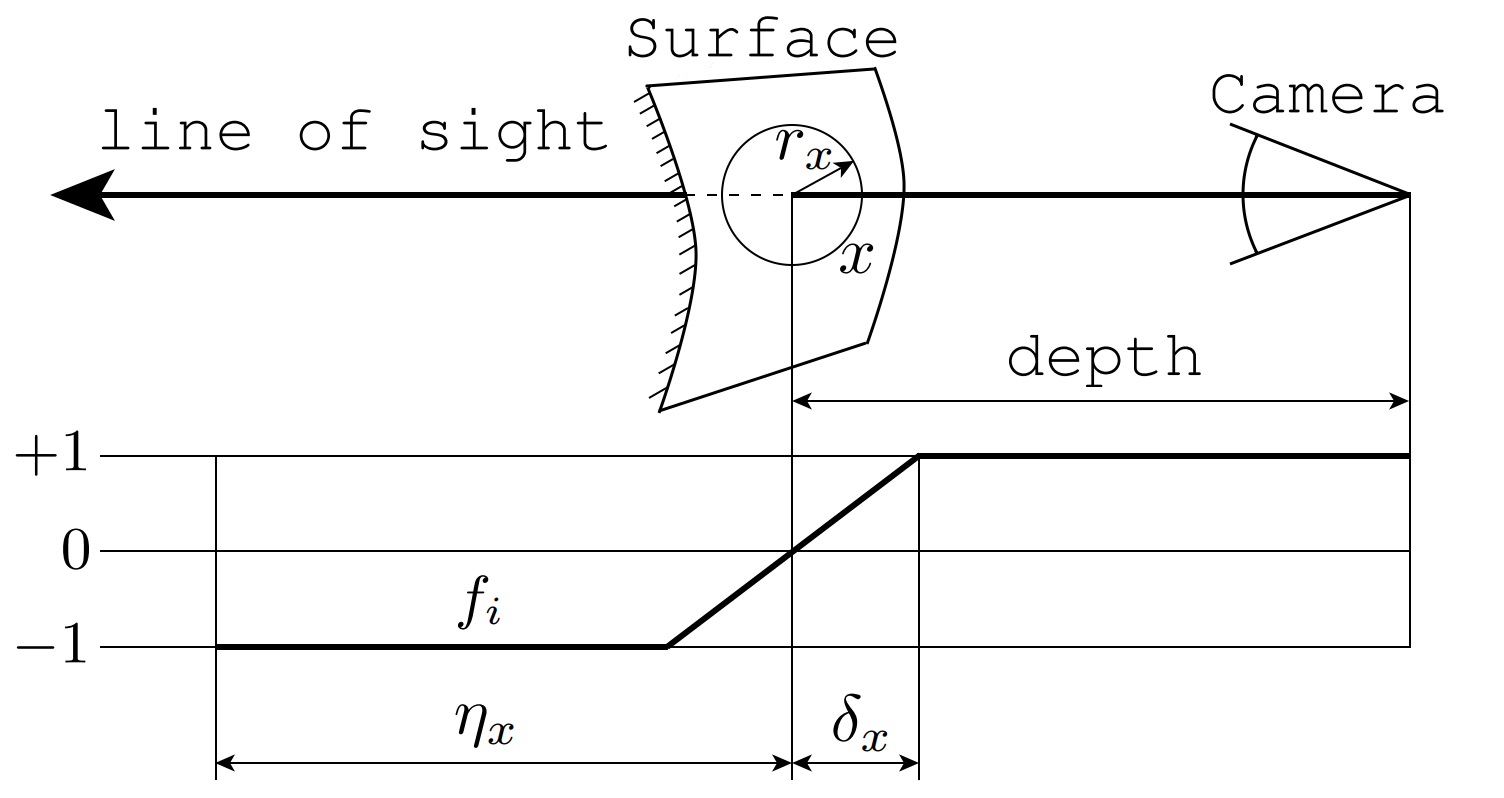}
    \caption{Generalization of a range field like in \cite{zach2007globally}. $f_i$ equal to $+1$ on each ray between the camera and a depth map sample and then fades away to $-1$ right under the surface sample.}
    \label{fig:range_field}
    \end{minipage}
    \begin{minipage}[b]{0.5\linewidth}
    \end{minipage}
    \begin{minipage}[b]{0.40\linewidth}
    \capstart
    \includegraphics[width=\textwidth]{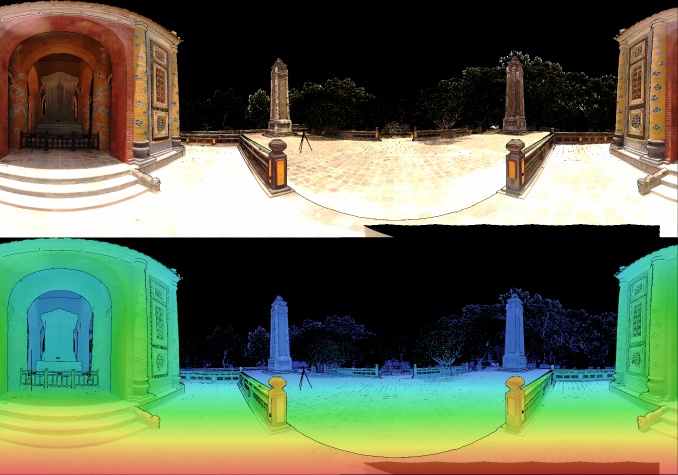}
    \caption{An example of a terrestrial LIDAR scan from the Tomb of Tu Duc dataset (RGB colors and depth from the sensor).}
    \label{fig:lidar_360_example}
    \end{minipage}
\end{figure}

It means that we need to generalize over all these types of data and work with an abstraction of a range image that can be used in the functional formulation.
All this data can naturally be formulated in a way, described in \cite{zach2007globally},
as a distance field $f_i$, which is equal to $+1$ on each ray between the camera and a depth map sample and then fades away to $-1$ right under the surface sample -- see Fig.~\ref{fig:range_field}.
The only difference for our case is that we want to work with scenes with diverse scales, so $\delta$ (the width of a relevant near-surface region) and $\eta$ (the width of an occluded region behind the surface) should be adaptive to the radius $r_x$ of a material point $x$.
Thus, in all our experiments we use $\delta_x=6 \cdot r_x$ and $\eta_x=3 \cdot \delta_x=18 \cdot r_x$ (smaller values lead to holes in thin surfaces, larger values lead to 'bubbliness' - excessive thickness of thin surfaces).

Both depth maps from stereo photos and RGB-D cameras can be represented in such way naturally.
The only non-trivial question is how to estimate material point radii $r_x$ for each pixel in a depth map.
For each depth map's pixel, we are estimating the distance in 3D space to the neighboring pixels and take half of that distance as the sample point's radius $r_x$.

To represent terrestrial LIDAR point clouds as range images, we rely on the fact that the structure of such point clouds is very similar to the structure of pictures, taken with 360-degree cameras (see Fig.~\ref{fig:lidar_360_example}).
Because of that, we treat them just like depth maps of 360-degree cameras with the only difference that LIDAR data is nearly noise-free, and
thus we can rely on such data with more confidence (i.e. with a weaker regularization term) -- see Section \ref{sec:dist-fields} below.

\subsection{Functional Formulation}

In our task, given multiple distance fields $f_i$, we want to find such an indicator field $u$
(where $u=0$ corresponds to a reconstructed isosurface, $u=+1$ -- to the exterior of the object, and  $u=-1$ -- to the interior of the object)
that will closely represent these distance fields in some way.
One of the ways to formulate what would be a good field $u$ is to introduce some energy functional. The less energy the functional produces -- the better the indicator field is.

Total variation (TV) regularization force term for $u$ with an $L^1$ data fidelity term between $u$ and $f_i$ is one such energy functional named TV$L^1$ \cite{zach2007globally} and is defined as:

\begin{align}
    \min_{u} \left\{
        \int_{\Omega} \Big(
            |\nabla u| + \lambda \sum_{i} |u - f_i|
        \Big) dx
    \right\}.\;
    \label{fig:tvl1}
\end{align}

Note that while the TV term prevents the surface from having discontinuities, there is no term that would force a regularity of surface normals
to tend the reconstruction to piecewise polynomial functions of an arbitrary order (see details in \cite{pock2011tgv}).
Such term was introduced as a part of the $TGV$ energy functional via an additional vector field $v$ in \cite{bredies2010total} and it was adapted to 2.5D reconstruction in \cite{pock2011tgv}:

\begin{align}
    \min_{u, v} \left\{
        \int_{\Omega} \Big(
            \alpha_1 |\nabla u - v| + \alpha_0 |\mathcal{E} (v)| + \sum_{i} |u - f_i|
        \Big) dx
    \right\},\;
    \label{fig:tgv}
\end{align}

where $\mathcal{E} (v)$ denotes the symmetric gradient operator

\begin{align}
    \mathcal{E} (v) = \frac{\nabla v + \nabla v^T}{2}.
    \label{fig:tgv_e}
\end{align}

In order to minimize this $TGV$ functional, we use the primal-dual method \cite{pock2011tgv}.
Also, like in \cite{zach2007globally}, we have implemented primal-dual iterations over a coarse-to-fine scheme with the execution of iterations accelerated on GPU for a faster convergence.
We have found that $200$ iterations are enough for convergence on each level of the scheme.

\begin{figure*}
    \centering
    \capstart
    \begin{minipage}[b]{0.67\linewidth}
        \includegraphics[width=\textwidth]{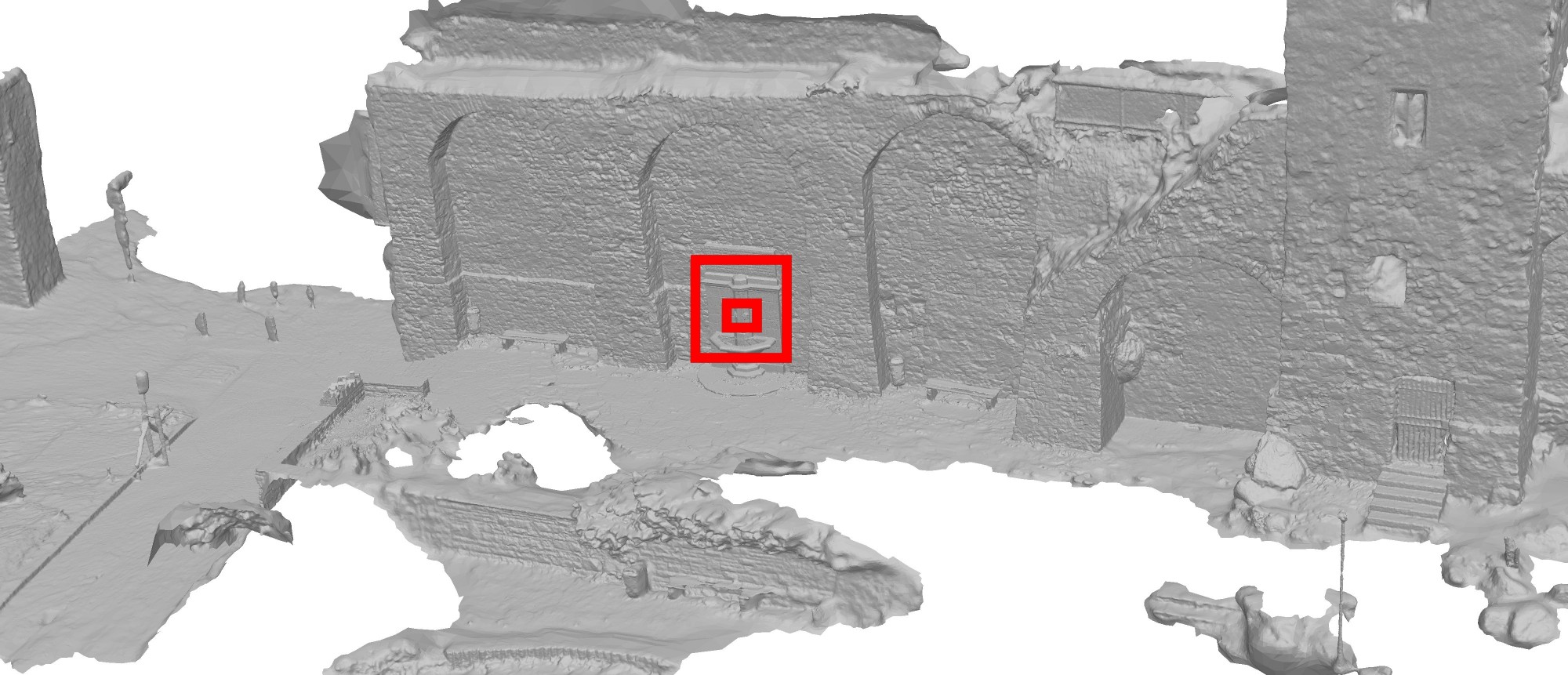}
        \begin{minipage}[b]{\linewidth}
            \begin{flushright}
            \begin{minipage}[b]{0.49\linewidth}
                \begin{includegraphics}[width=\textwidth]{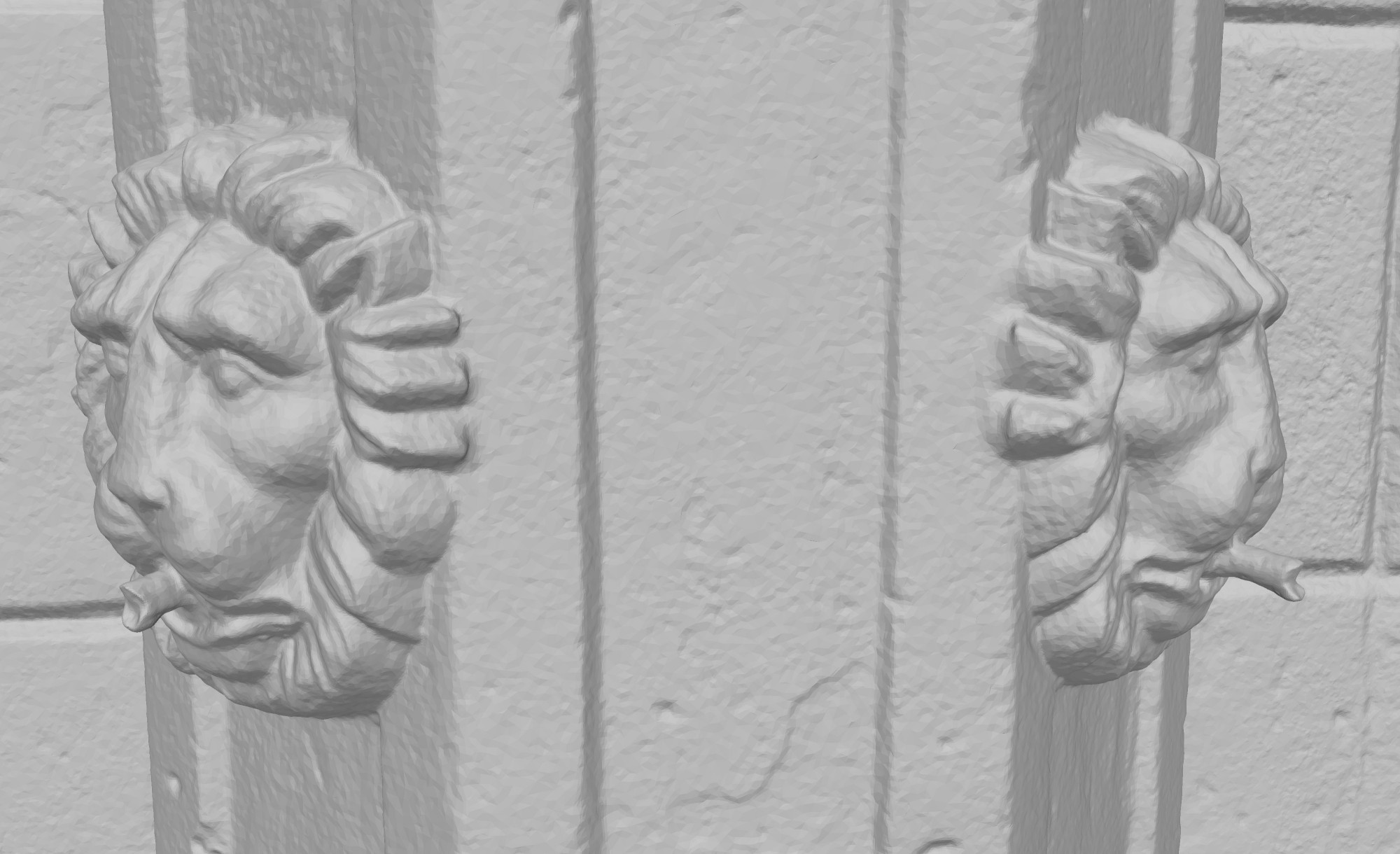}
                    \put(-160,87){Ours}
                \end{includegraphics}
            \end{minipage}
            \begin{minipage}[b]{0.49\linewidth}
                \begin{includegraphics}[width=\textwidth]{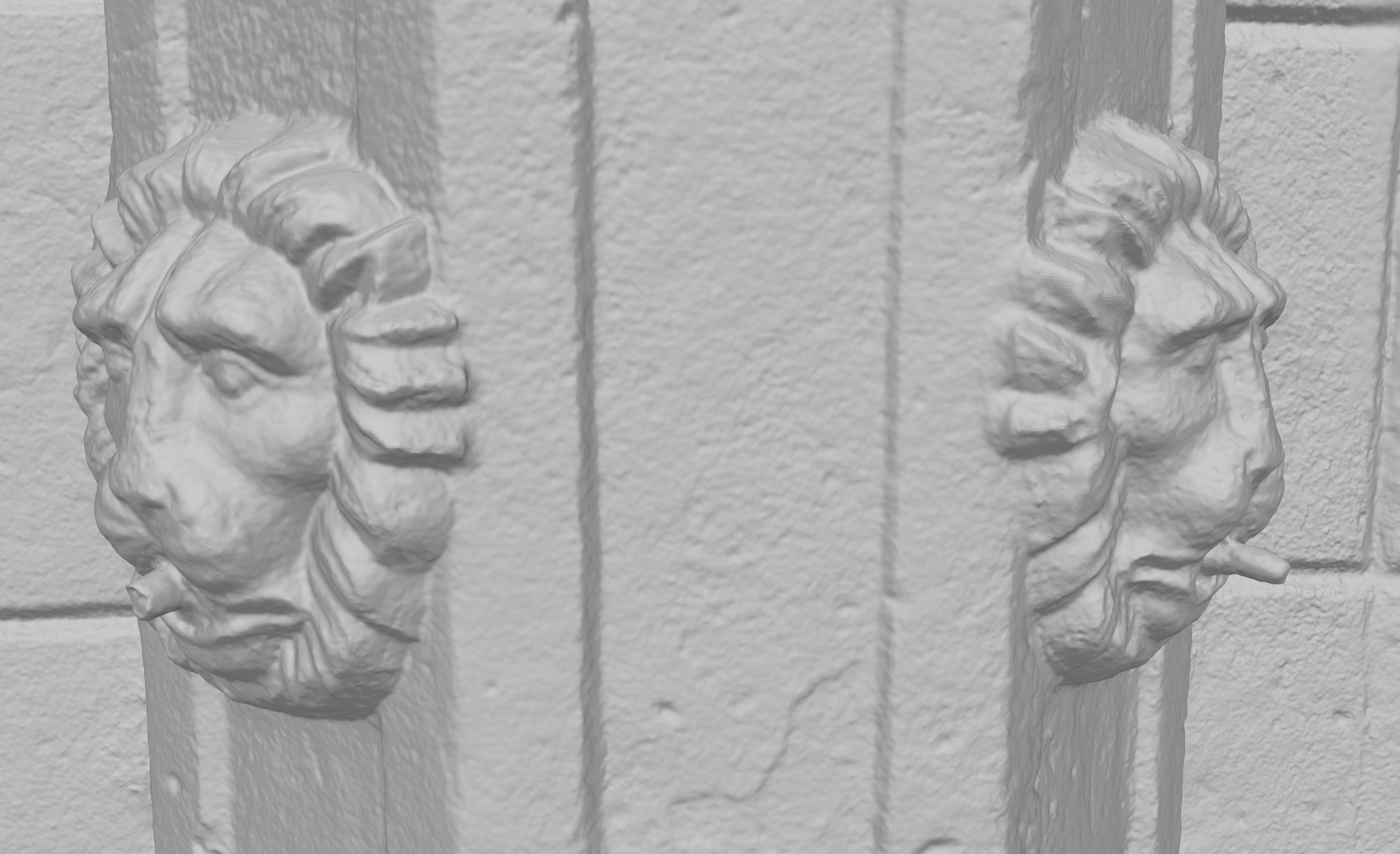}
                    \put(-160,87){GDMR}
                \end{includegraphics}
            \end{minipage}
            \end{flushright}
        \end{minipage}
    \end{minipage}
    \begin{minipage}[b]{0.31\linewidth}
        \begin{minipage}[b]{\linewidth}
            \begin{includegraphics}[width=\textwidth]{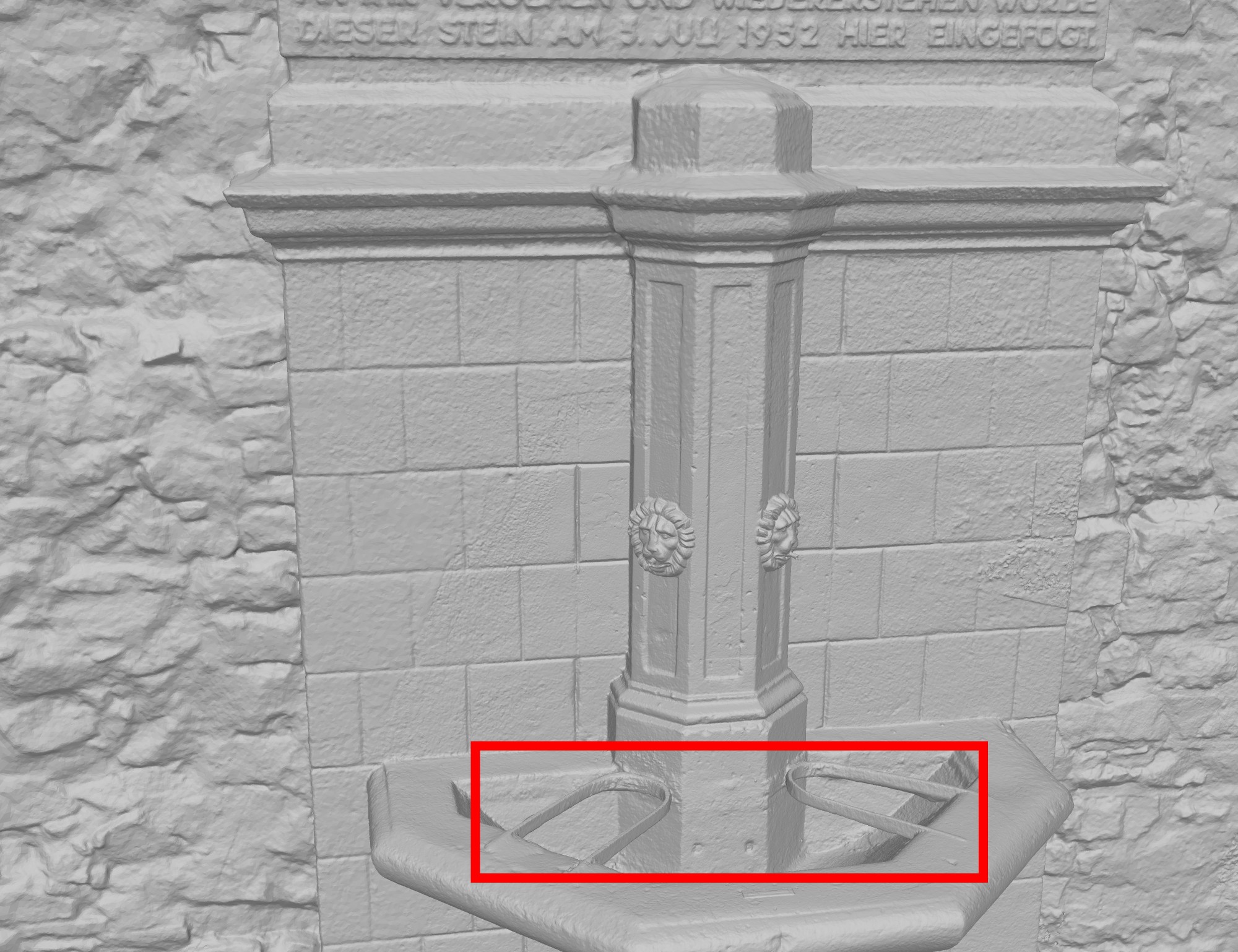}
                \put(-150,105){Ours}
            \end{includegraphics}
        \end{minipage}
        \begin{minipage}[b]{\linewidth}
            \begin{includegraphics}[width=\textwidth]{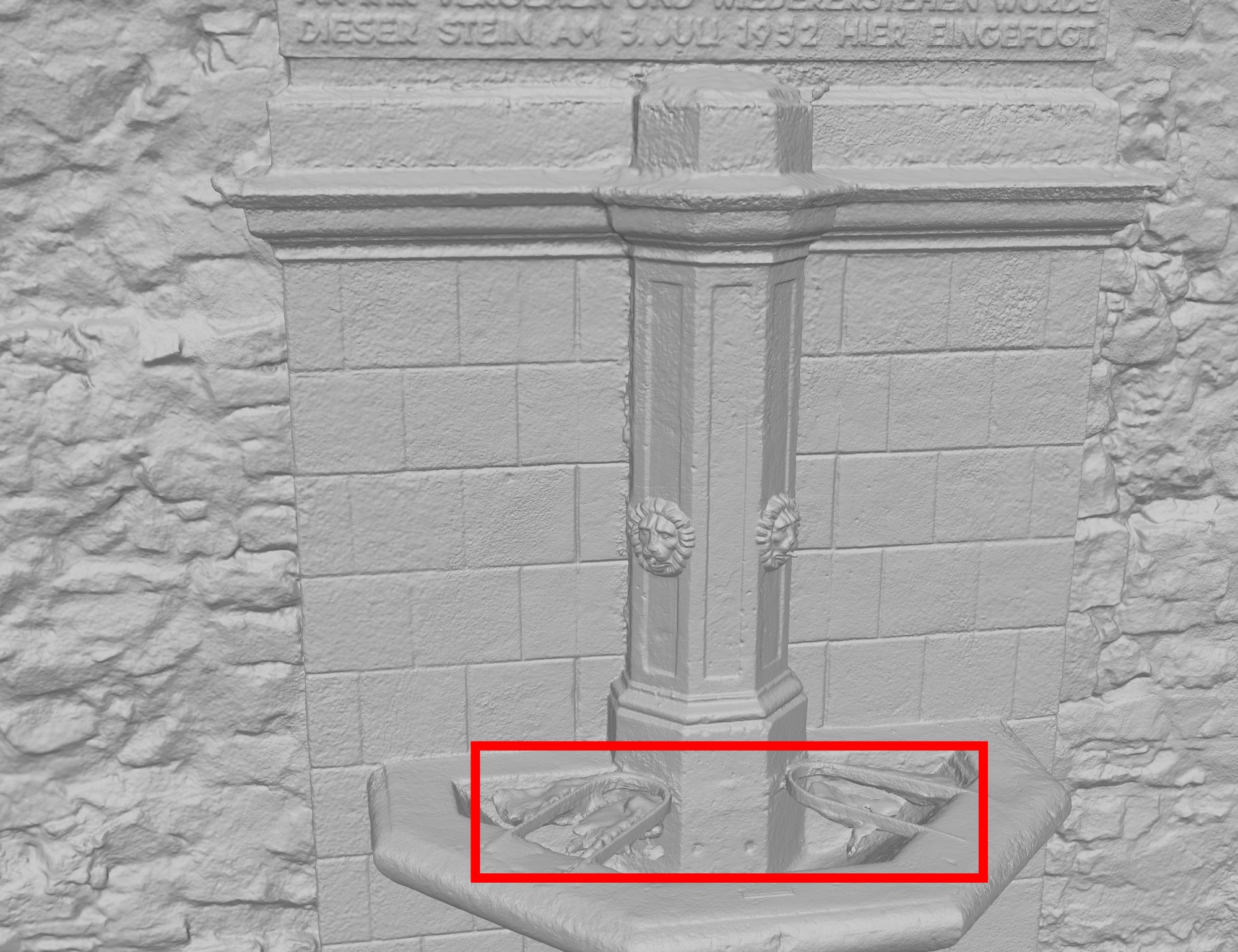}
                \put(-150,105){GDMR}
            \end{includegraphics}
        \end{minipage}
    \end{minipage}
    \caption{Results on the Citywall dataset and comparison with GDMR \cite{ummenhofer2015global} results.
    The results are comparable.
    Note that due to strong visibility-based noise-filtering properties our method led to the cleaner basin of the fountain.}
    \label{fig:citywall}
\end{figure*}

\subsection{Space Discretization}
\label{sec:space-repr}

To minimize $TGV$ w.r.t. $u$, we need to choose a space discretization. A regular grid \cite{zach2007globally} does not correspond to scale diversity and will potentially lead to high memory consumption. In our approach, we use an adaptive octree.
Let $r_{root}$ be the radius of the root cube of the octree.
For each point sample $x$ with the center in $p_x$ and the radius $r_x$, an octree should have
a cube $c$ containing $p_x$. This cube should be on such an octree depth $d_c$ that the following inequality holds:

\begin{align}
    0.75 \cdot r_x \le \frac{r_{root}}{2^{d_c}} < 1.5 \cdot r_x.
    \label{fig:cube_depth}
\end{align}

In the implemention of the primal-dual method of minimization, we need pointers from each octree cube to its
neighboring cubes in order to access neighbors' $u$, $v$ values and dual variables $p$, $q$. In an adaptive octree, any cube can have any number of neighbors
due to adaptive subdivision and scene scale diversity. However, we aim to execute the iterative scheme on GPU, meaning that in order
to achieve good performance we need to limit the number of neighboring voxels in some way, resulting in a limited number of references needed to store per each voxel.
We utilize the approach, discussed in\cite{ummenhofer2015global}, which results in 2:1 balancing of the adaptive octree, leading to the point when each cube has only $4$ or less neighbors over each face.

As will be discussed later in Section \ref{sec:dist-fields}, it is important to know what average radius of point samples $S_c$ corresponds to each octree cube $c$ creation. Therefore,
for each octree cube we also store its density, which we also call the cube's radius $r_c$, defined as

\begin{align}
    r_c = \frac{ \sum_{x \in S_c} r_x}{| S_c |}.
    \label{fig:cube_radius}
\end{align}

\subsection{Distance Fields to Histograms}
\label{sec:dist-fields}

We transform all distance fields into histograms in our octree like in \cite{zach2007globally}. This allows us
to run iterations with compact histograms \cite{zach2008fast} (thanks to fixed size per voxel) instead of large (due to high overlap) depthmaps \cite{zach2007globally} in memory.
The main difference is that we want the algorithm to be aware of scale diversity and to implement the minimization framework in the coarse-to-fine scheme. Because of this, it is impossible to ignore how big or small a voxel projection is in a depth map: if a projected voxel is large (for example, on the coarser levels) and is covered with many depth map pixels (i.e. intersected with many distance field rays), we need to account for all of them. This problem is very similar to the texture aliasing problem in computer graphics, which can be solved
with texture mipmaps \cite{williams1983pyramidal}. Similarly, we have used depth map pyramids by building mipmaps for each depth map. Therefore, when we
project a voxel to the depth map, we choose an appropriate depth map level of details, and only then we estimate a histogram bin of a current voxel to which the current depth map will contribute,
see the listing in Algorithm \ref{alg:hists_listing}.

Note that such voxel projection into the pyramid of a depth map is very unstable and changes heavily with any change of the size of a working region bounding box, which happens because the natural voxel's radius in the octree is equal to $\frac{r_{root}}{2^{d}}$, where $d$ is the voxel's depth and $r_{root}$ is
the radius of the octree's root voxel which depends on the size of a whole working region. To be invariant to the selection of the working region size,
for each cube, in addition to its center, we store its density value, which is equal to the average radius $r_c$ of point samples that it represents, see Eq.~\ref{fig:cube_radius}.

\begin{figure}
    \centering
    \capstart
    \begin{minipage}[b]{\linewidth}
        \begin{minipage}[b]{0.49\linewidth}
            \includegraphics[width=\textwidth]{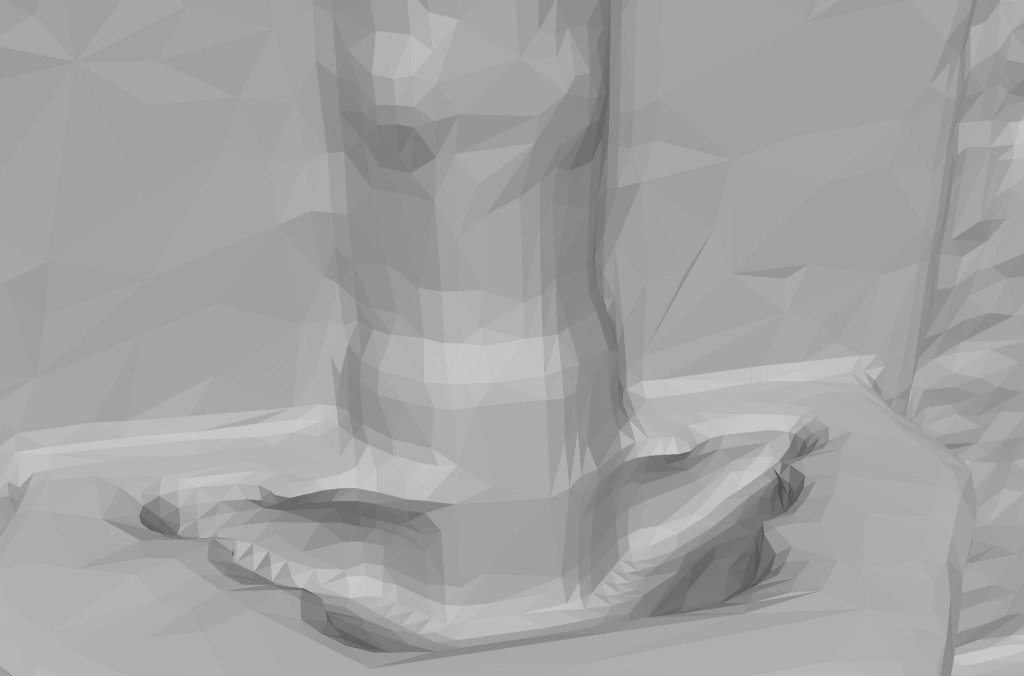}
        \end{minipage}
        \begin{minipage}[b]{0.49\linewidth}
            \includegraphics[width=\textwidth]{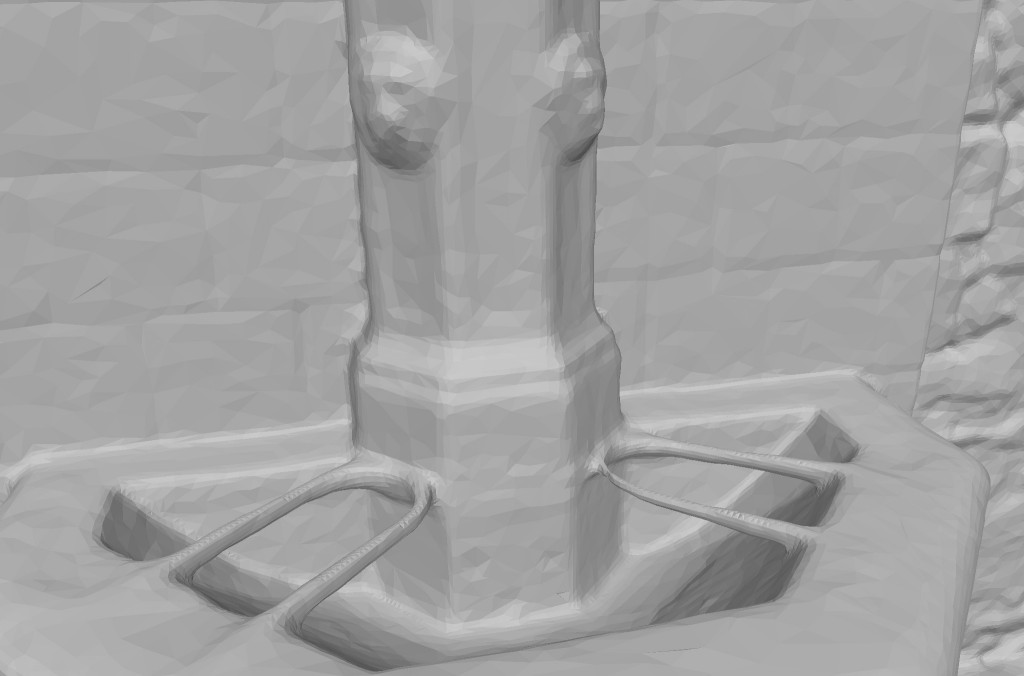}
        \end{minipage}

        \begin{minipage}[b]{0.49\linewidth}
            \includegraphics[width=\textwidth]{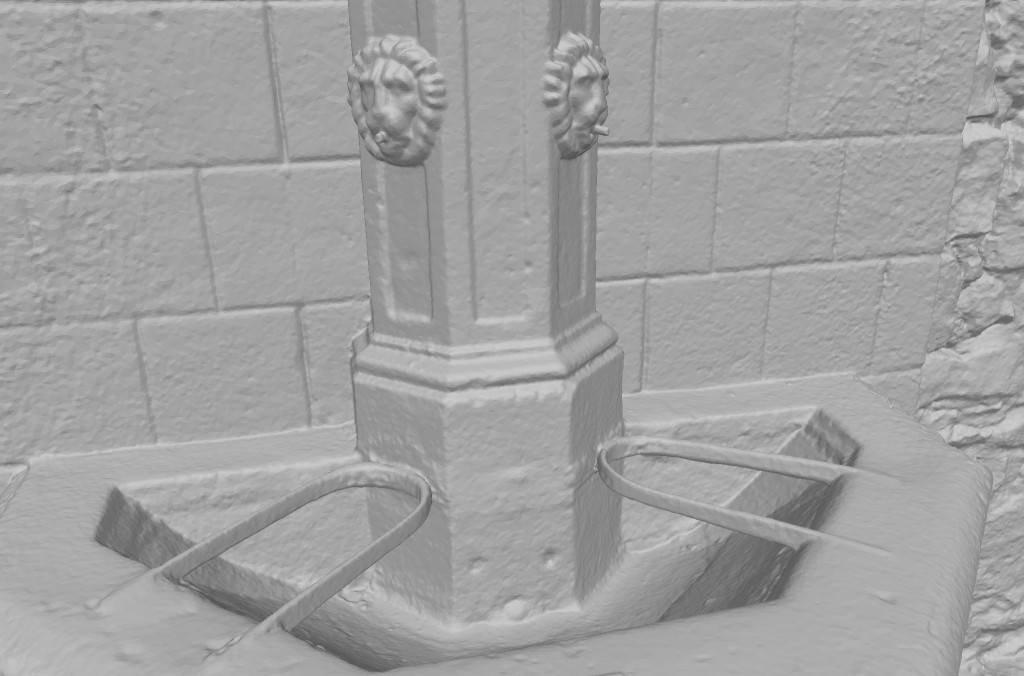}
        \end{minipage}
        \begin{minipage}[b]{0.49\linewidth}
            \includegraphics[width=\textwidth]{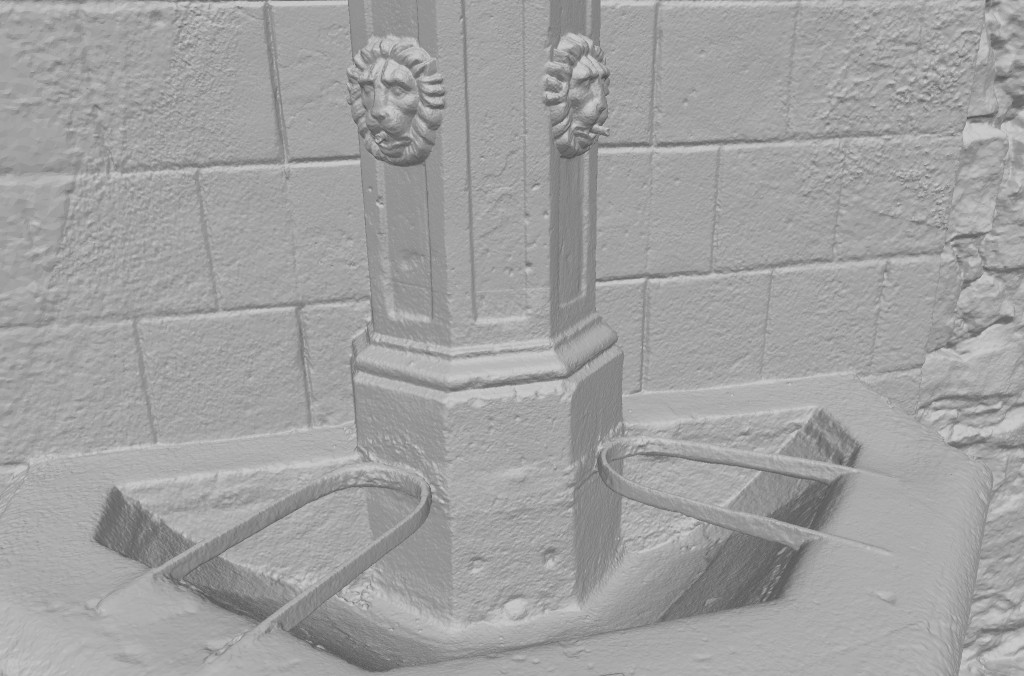}
        \end{minipage}
    \end{minipage}
    \caption{Citywall dataset. Isosurface progression from coarse to fine levels: 11, 13, 15, 17.
    Due to cube radius robustness described in Section \ref{sec:dist-fields}, we see local progressive isosurface changes
    from level to level. Also, note the progressive refinement of topology, which can not be achieved in photoconsistent refinement methods \cite{vu2011high}, \cite{li2016efficient}
    because they refine polygonal surfaces directly (with a 1-to-4 triangle subdivision and vertex movement). Instead, we refine the indicator field, so the topology
    of our implicit isosurface corresponding to zero indicator value changes together with the indicator field.}
    \label{fig:citywall_coarse_to_fine}
\end{figure}

These details lead to local and stable progressive isosurface refinement from coarse to fine levels - see Fig.~\ref{fig:citywall_coarse_to_fine}.

\begin{algorithm*}
\capstart
\caption{Pseudocode of the estimation of distance field contribution to voxel histograms.}
\label{alg:hists_listing}
\begin{algorithmic}[H]
\Procedure{add\_to\_voxel\_histograms}{$\vars{depthmap\_pyramid}, \vars{voxel}$}
    \State $\vars{mipmap\_level}, \vars{pixel} \gets \vars{depthmap\_pyramid}.\func{project}(\vars{voxel.center}, \vars{voxel.radius})$
    \State $\vars{depthmap} \gets \vars{depthmap\_pyramid}.\func{get}(\vars{mipmap\_level})$
    \State $\vars{depth} \gets \vars{depthmap}.\func{get}(\vars{pixel})$
    \If{$\vars{depth} = \vars{None}$}
        \State \Return \Comment{If the distance field doesn't have ray in such direction -- it doesn't contribute to such voxel anything.}
    \EndIf
    \State $r_x \gets \vars{voxel.radius}$ \Comment{See Section \ref{sec:dist-fields-as-input-data} and Fig.~\ref{fig:range_field}:}
    \State{$\delta_x \gets 6 \cdot r_x$} \Comment{$\delta_x$ -- width of the relevant near-surface}
    \State{$\eta_x \gets 3 \cdot \delta_x$} \Comment{$\eta_x$ -- width of the occluded region.}
    \State{$\vars{distance} \gets \vars{depthmap\_pyramid}.\func{distance\_to}(\vars{voxel.center})$}
    \State{$a \gets \vars{depth - distance}$} \Comment{$a$ equals to zero if the voxel is exactly at the observed surface level.}
    \If{$a < -\eta_x$} \Comment{If the voxel is further then occluded region behind the surface observed with ray then we do not}
        \State \Return \Comment{observe such voxel from current depth map, i.e. it does not contribute to the voxel's histograms.}
    \EndIf
    \State $\vars{depthmap\_vote} \gets 1$ \Comment{Or $\vars{depthmap\_vote} \gets 5$ in case of noise-free terrestrial LIDAR input.}
    \State ${a} \gets \func{max}(-1.0, \func{min}(1.0, a / \delta_x))$ \Comment{Clamp to the indicator range.}
    \State $\vars{bin} \gets \func{floor}(((a + 1.0) / 2.0) \cdot 8.0)$ \Comment{We are using 8 bins following \cite{zach2008fast}.}
    \State $\vars{voxel.histograms[bin]} \gets \vars{voxel.histograms[bin]} + \vars{depthmap\_vote}$
\EndProcedure
%
%
\end{algorithmic}
\end{algorithm*}

\section{Out-of-Core Adaptation}
\label{sec:out-of-core}

Our main contribution is an out-of-core adaptation of the $TGV$ minimization scheme on a 2:1 balanced octree. Consequently,
we implement each stage of the algorithm in an out-of-core way, where the stages are:

\begin{itemize}
    \item[\textbf{\ref{sec:linear-octree}}] Build a linear octree from all cubes discussed above in Space discretization (Section \ref{sec:space-repr})
    \item[\textbf{\ref{sec:balanced-octree}}] Balance the octree, so that each cube has a limited number of neighbors
    \item[\textbf{\ref{sec:treetop}}] Build an indexed treetop to run primal-dual iterations independently on each treetop leaf's part
    \item[\textbf{\ref{sec:hists-init}}] Save distance fields' votes to voxels' histogram bins over the balanced octree (GPU-accelerated)
    \item[\textbf{\ref{sec:iters}}] Coarse-to-fine functional minimization over each part of the balanced octree (GPU-accelerated)
    \item[\textbf{\ref{sec:cubes-marching}}] Surface extraction via the marching cubes algorithm
\end{itemize}

\subsection{From Distance Fields to Octree}
\label{sec:linear-octree}

For each distance field, we estimate each sample's radius $r_x$ as half of the distance to its neighbors in this distance field.
Then for each sample, we spawn an octree cube containing this sample at an appropriate to $r_x$ depth $d$, as formulated in Eq.~\ref{fig:cube_depth}.
All cubes are encoded with 96-bit 3D Morton codes \cite{morton1966computer} and are saved to a single file per each distance field.

Afterwards, we need to merge all these files containing cubes (i.e. Morton codes). We encoded our cubes with Morton codes, which introduce a Z-curve order over them -- see a Z-curve example on a balanced 2D quadtree on Fig.~\ref{fig:zcurve}.
Thus, cubes merging into a single linear octree can be done with an out-of-core k-way merge sort.

\subsection{Octree Balancing}
\label{sec:balanced-octree}

To limit the number of neighbors for each cube, we need to balance the obtained octree. A linear octree can be arbitrarily large because
it describes the whole scene. Out-of-core octree balancing is described in \cite{tu2004balance}.
Balancing also relies on the Morton code ordering -- we only need to load a part of the sorted linear octree, balance that part independently from others,
and save the balanced part to a separate file.
Later, we only need to merge all balanced parts, which can be accomplished like in the previous stage of linear octree merging -- via an out-of-core k-way merge sort.

\subsection{Octree Treetop}
\label{sec:treetop}

\begin{figure}
    \centering
    \begin{minipage}[b]{0.45\linewidth}
    \capstart
    \begin{includegraphics}[width=\textwidth]{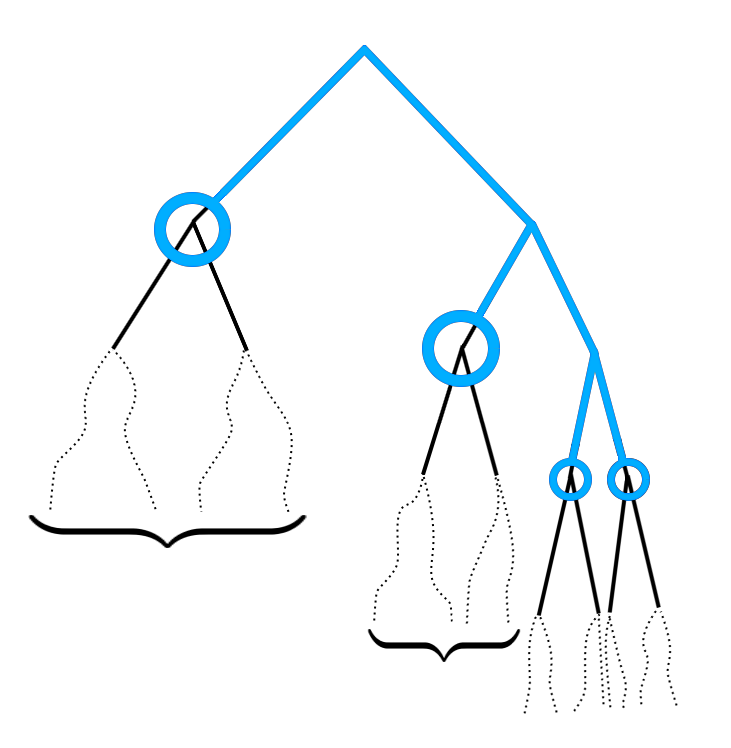}
        \put(-85,29){\rotatebox{-90}{$<$}}
        \put(-87,11){{$N$}}
        \put(-45,12){\rotatebox{-90}{$<$}}
        \put(-47,-4){{$N$}}
        \put(-30,7){\rotatebox{-90}{$\}<$}}
        \put(-30,-15){{$N$}}
        \put(-18,7){\rotatebox{-90}{$\}<$}}
        \put(-18,-15){{$N$}}
    \end{includegraphics}
    \caption{Blue nodes are the treetop leaves with less than $N_{cubesPerTreetopLeaf}$ cubes under each subtree.}
    \label{fig:treetop}
    \end{minipage}
    \begin{minipage}[b]{0.10\linewidth}
    \end{minipage}
    \begin{minipage}[b]{0.45\linewidth}
    \capstart
    \includegraphics[width=\textwidth]{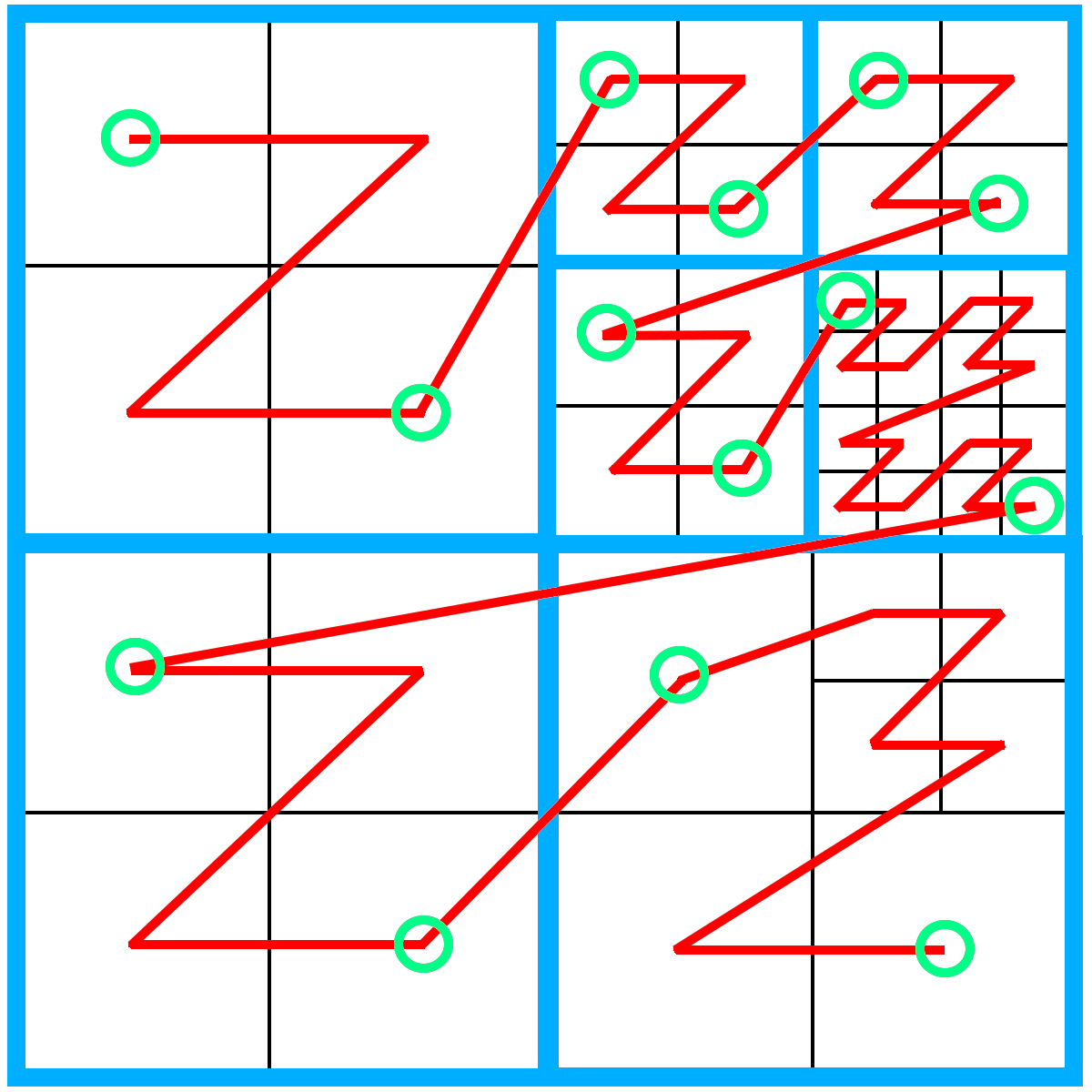}
    \caption{Note that the Z-curve enters and leaves cubes (green circles) from each treetop leaf (blue boxes) exactly once.}
    \label{fig:zcurve}
    \end{minipage}
\end{figure}

At this moment, we need to have some high-level scene representation to be able to compute the histograms and
run iterations for $TGV$ minimization over the balanced octree part by part. In fact, such
subdivision into parts will make it possible for each of the next stages, including the final polygonal surface extraction,
to be split into OpenCL's $workItem$-like independent parts (i.e. with massive parallelism, which is useful for cluster-acceleration).

Let us calculate how many descendants each intermediate cube of the octree has on deeper octree levels.
Consider a treetop -- a rooted subtree that contains the minimum number of octree cubes in the leaves, with the restriction that each
leaf cube contains less than $N_{cubesPerTreetopLeaf}$ descendants in the original tree, see Fig.~\ref{fig:treetop}.
In all experiments we used $N_{cubesPerTreetopLeaf} = 2^{24}$
because it is small enough to guarantee that each subsequent step will fit in 16 GB of RAM, but at the same time limits
the number of leaves in a treetop to just a couple of thousands even on the largest datasets.

Due to out-of-core constraints, we can not estimate a global treetop by loading the whole octree into the memory.
Therefore, we build independent treetops for all linear balanced octree parts, and then merge those treetops
into a global one. At this stage, we can easily save indices of all covered relevant cubes for each treetop leaf.
Moreover, these indices are consecutive due to Z-curve ordering of Morton codes -- see Fig.~\ref{fig:zcurve}.
Hence, we only need to save two indices with each treetop leaf --indices of the first and the last relevant cubes from the linear balanced octree.
This gives us an ability to load cubes relevant for current treetop leaf from balanced octree in IO-friendly consecutive way.
In addition, we have strong guarantees that the number of such cubes is limited by $N_{cubesPerTreetopLeaf}$.

\subsection{Computation of the histograms}
\label{sec:hists-init}

Now we have the scene representation, provided by the balanced linear octree and its indexed treetop. As the next part of our method,
we need to add votes of all distance fields to all relevant cubes in the octree.

Let us process all treetop leafs one by one and estimate relevant distance fields for each leaf, which is achieved simply by
checking each distance field frustum for an intersection with the treetop leaf cube volume. Then, we can just load all relevant distance fields
for each treatop leaf one by one and add their votes to all descendants
of the current treetop leaf, like shown in the listing in Algorithm \ref{alg:hists_listing}.

Note that at any moment during the computation of the histograms the memory contains no more than $N_{cubesPerTreetopLeaf}$ octree cubes and only
a single distance field.

\begin{figure}
    \centering
    \capstart
    \begin{minipage}[b]{0.68\linewidth}
        \includegraphics[width=\textwidth]{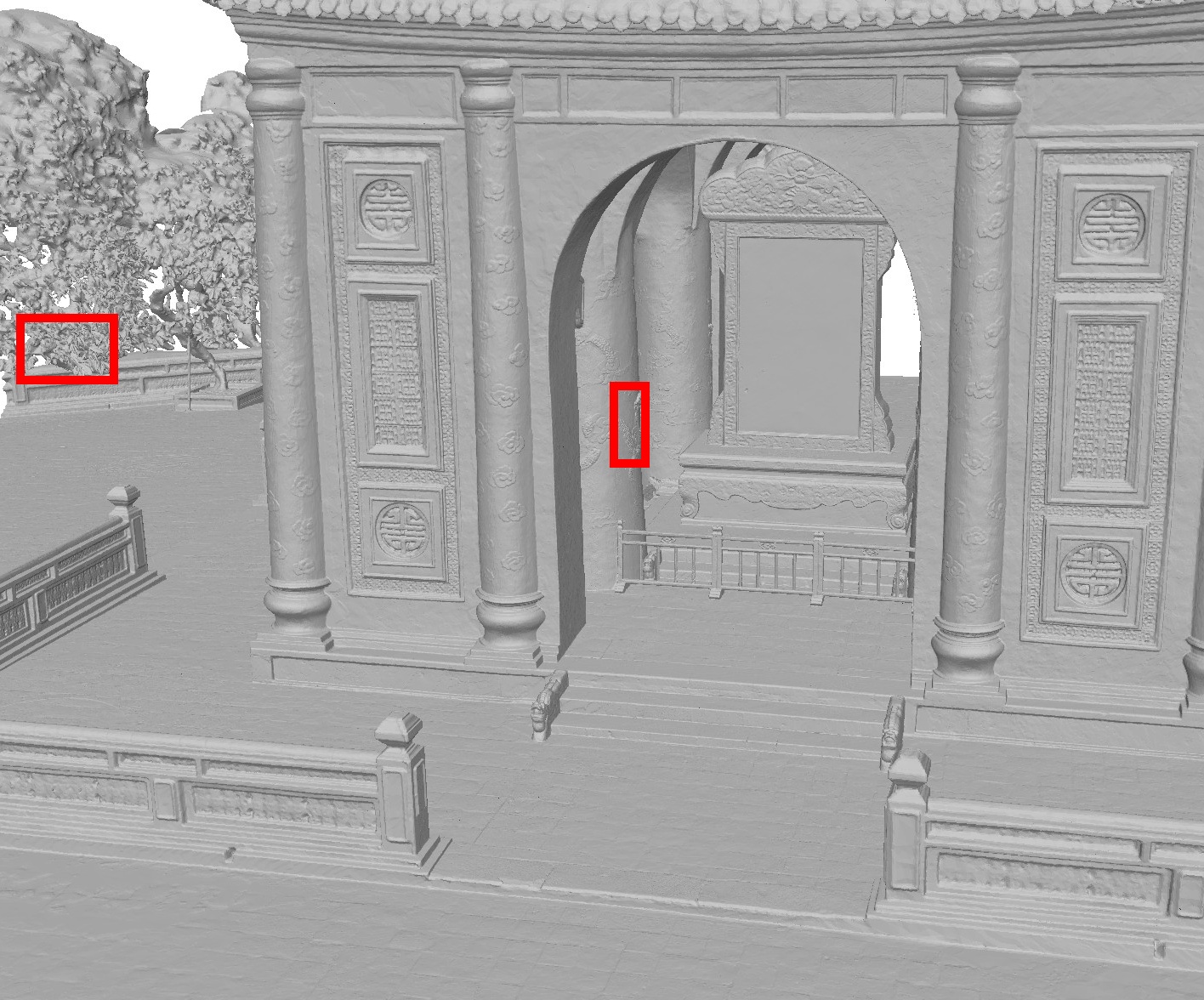}
    \end{minipage}
    \begin{minipage}[b]{0.30\linewidth}
        \begin{minipage}[b]{\linewidth}
            \includegraphics[width=\textwidth]{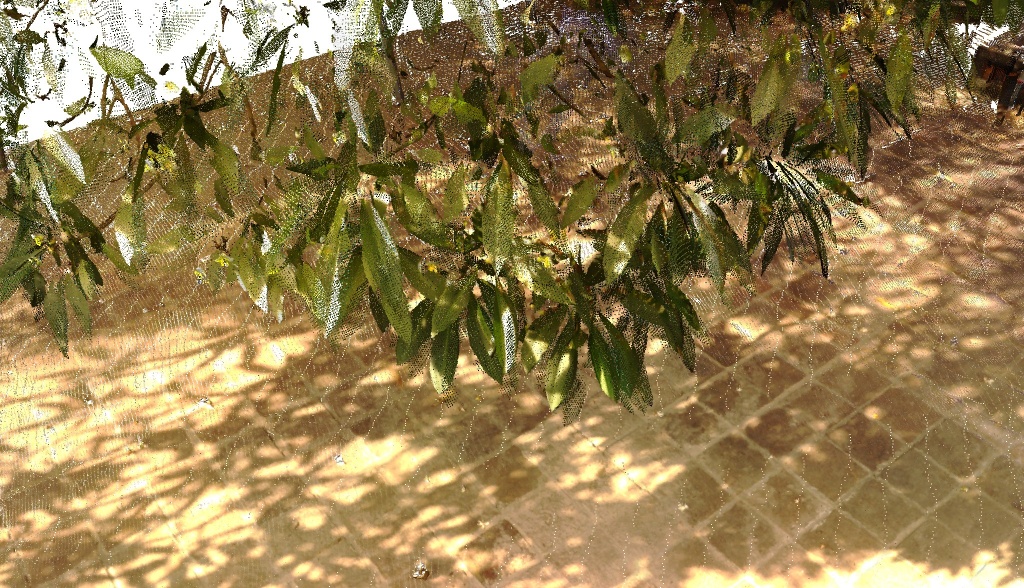}
        \end{minipage}
        \begin{minipage}[b]{\linewidth}
            \includegraphics[width=\textwidth]{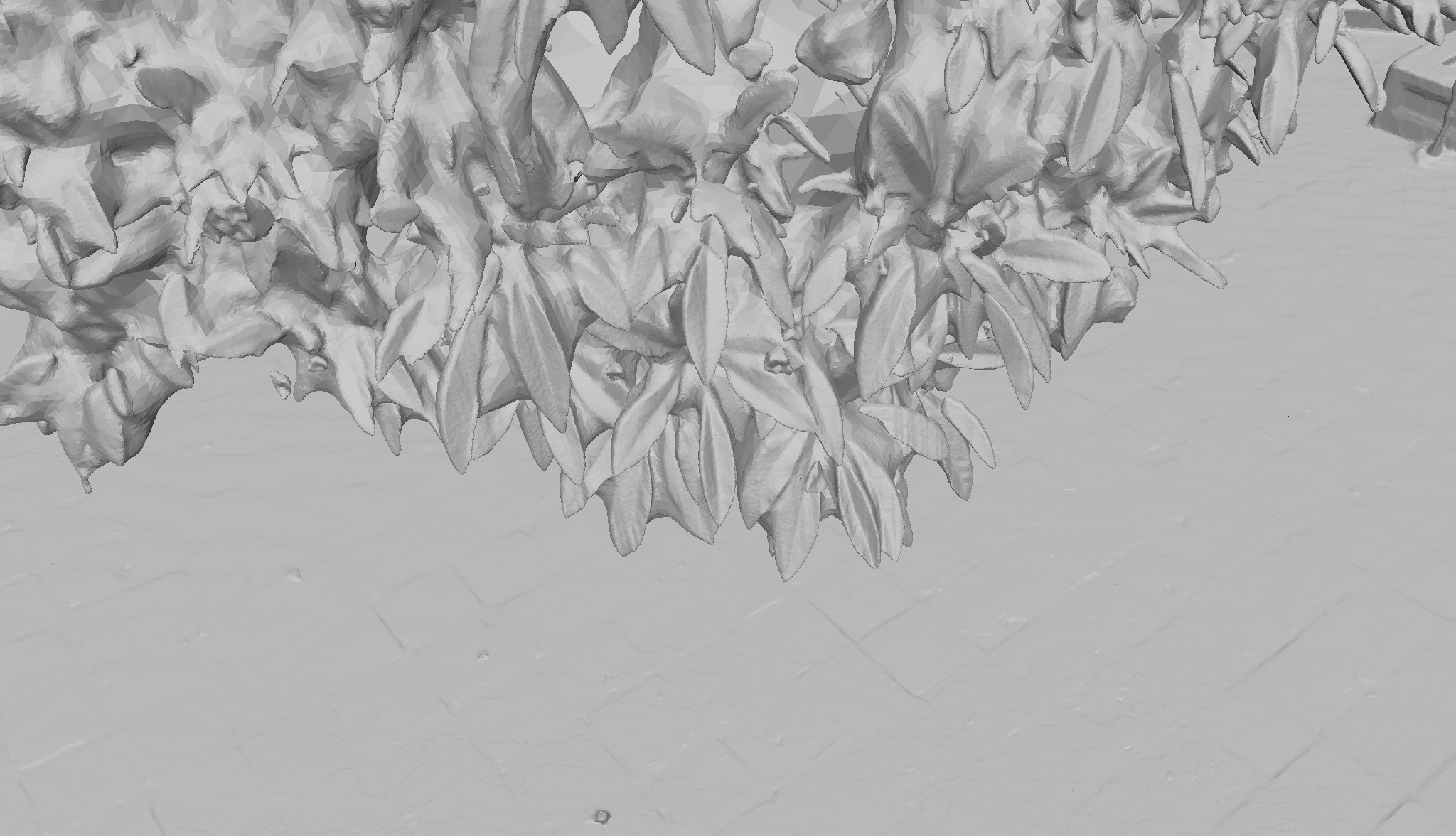}
        \end{minipage}
        \begin{minipage}[b]{0.48\linewidth}
            \includegraphics[width=\textwidth]{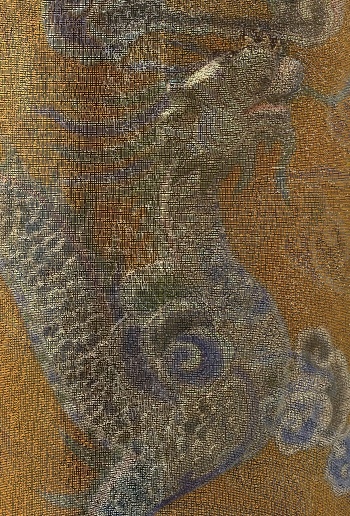}
        \end{minipage}
        \begin{minipage}[b]{0.48\linewidth}
            \includegraphics[width=\textwidth]{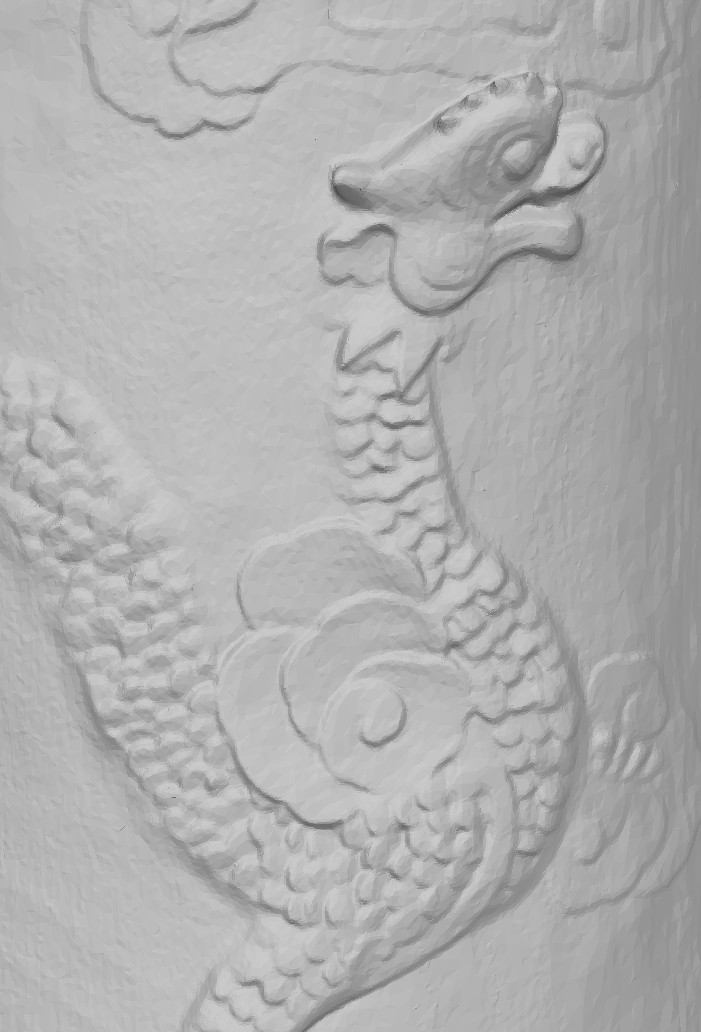}
        \end{minipage}
    \end{minipage}
    \caption{Tomb of Tu Duc LIDAR dataset. To the right -- two closeups with colored LIDAR point clouds and resulting models.}
    \label{fig:tomb_lidar}
\end{figure}

\subsection{Functional Minimization}
\label{sec:iters}

Now we need to iteratively minimize the $TGV$ functional from Eq.~\ref{fig:tgv}. As shown in \cite{zach2007globally}, it is
highly beneficial to use a coarse-to-fine scheme (especially in regions with lack of data) for convergence speed.
As we will see in this subsection -- the scheme also helps to not introduce any seams between processed parts.

\begin{figure}
    \begin{minipage}[b]{\linewidth}
        \centering
        \capstart
        \includegraphics[width=0.9\textwidth]{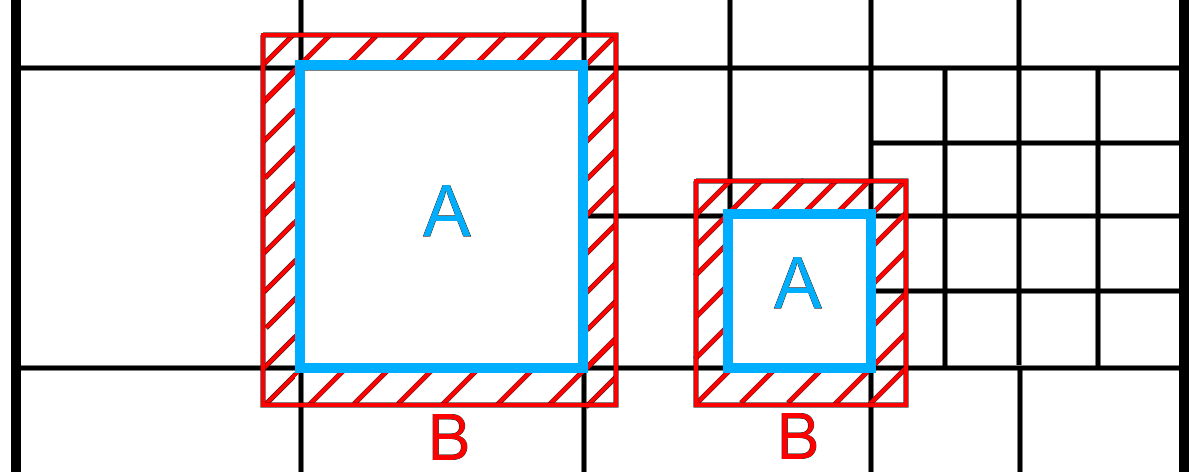}
        \caption{To avoid having visible seams we update the indicator for all cubes inside the current leaf's border (set $A$)
            while the indicator for neighboring cubes outside of the border (set $B$) is frozen.}
        \label{fig:fixed_indicators}
    \end{minipage}
\end{figure}

Suppose that we have already minimized the functional over the whole octree up to depth $level$. Now we want to execute
primal-dual iterations at the depth $level+1$ in an out-of-core way while not producing any seams between the parts. Like in the previous stage during the computation of the histograms,
we process treetop leaves one by one. Let us load a treetop leaf's cubes in a set $A$ and their neighbors
in a set $B$. Now, we can iterate a numeric scheme like in \cite{pock2011tgv} over the cubes from $A$ with the only difference on the treetop leaf's border
-- we want our neighbors' indicator values $u$ in cubes from $B$ to be equal to indicator values of their parenting cubes,
which were estimated on the previous level thanks to the coarse-to-fine scheme.
I.e. we update the indicator for all cubes inside the current leaf's border (set $A$)
while the indicator for neighboring cubes outside of the border (set $B$) is frozen - see two examples in Fig.~\ref{fig:fixed_indicators}.

By following this routine, at any given time we process only the cubes from a treetop leaf and their neighbors, and thus our memory
consumption is bounded by their number. We do not face any misalignments on the surface next to treetop leaf borders due to explicit
border constraints and the fact that the surface from one level to the next does not move far, but just progressively becomes more detailed,
see Fig.~\ref{fig:citywall_coarse_to_fine} and details of the computation of the histograms in Section \ref{sec:dist-fields}.

We notice that not so many cubes appear on the coarsest levels, meaning that each separate treetop leaf normally contains very few cubes.
Therefore, we find it beneficial for performance to process multiple treetop leaves at once on the coarsest levels (w.r.t. leaves' total number of cubes on the current level).

\subsection{Marching Cubes}
\label{sec:cubes-marching}

As the last part, after estimating the indicator value $u$ for all cubes of the octree, we need to extract a polygonal iso-surface corresponding to an
indicator value $u=0$. For that purpose, we can perform the marching cubes algorithm on per-leaf basis, using the same out-of-core tree partition as used before.

Marching cubes in a part of balanced octree is trivial: for each octree cube, we extract 3D-points between indicator values of different
sign (i.e. points from iso-surface corresponding to zero indicator value), and then build their triangulation via dynamic programming
by minimizing the total surface area, similar to \cite{bloomenthal1988polygonization}.

Note that neighboring surface parts have seamlessly matching borders,
because both parts have the same indicator value $u$ across the border due to stability of progressive refinement
discussed in the previous subsection.

Finally, it is important to note that the number of triangle faces will be extra-large for any large dataset.
We follow each octree part marching cubes with QSlim-based \cite{garland1998simplifying} decimation,
which we modified with a strict border constraint, namely that no border edge (i.e. a triangle edge lying on a treetop leaf's cube face) should be ever collapsed.
This way we achieve strict guarantees of a seamless surface between neighboring treetop leaves.

\begin{figure}
    \centering
    \capstart
    \begin{minipage}[b]{0.49\linewidth}
        \includegraphics[width=\textwidth]{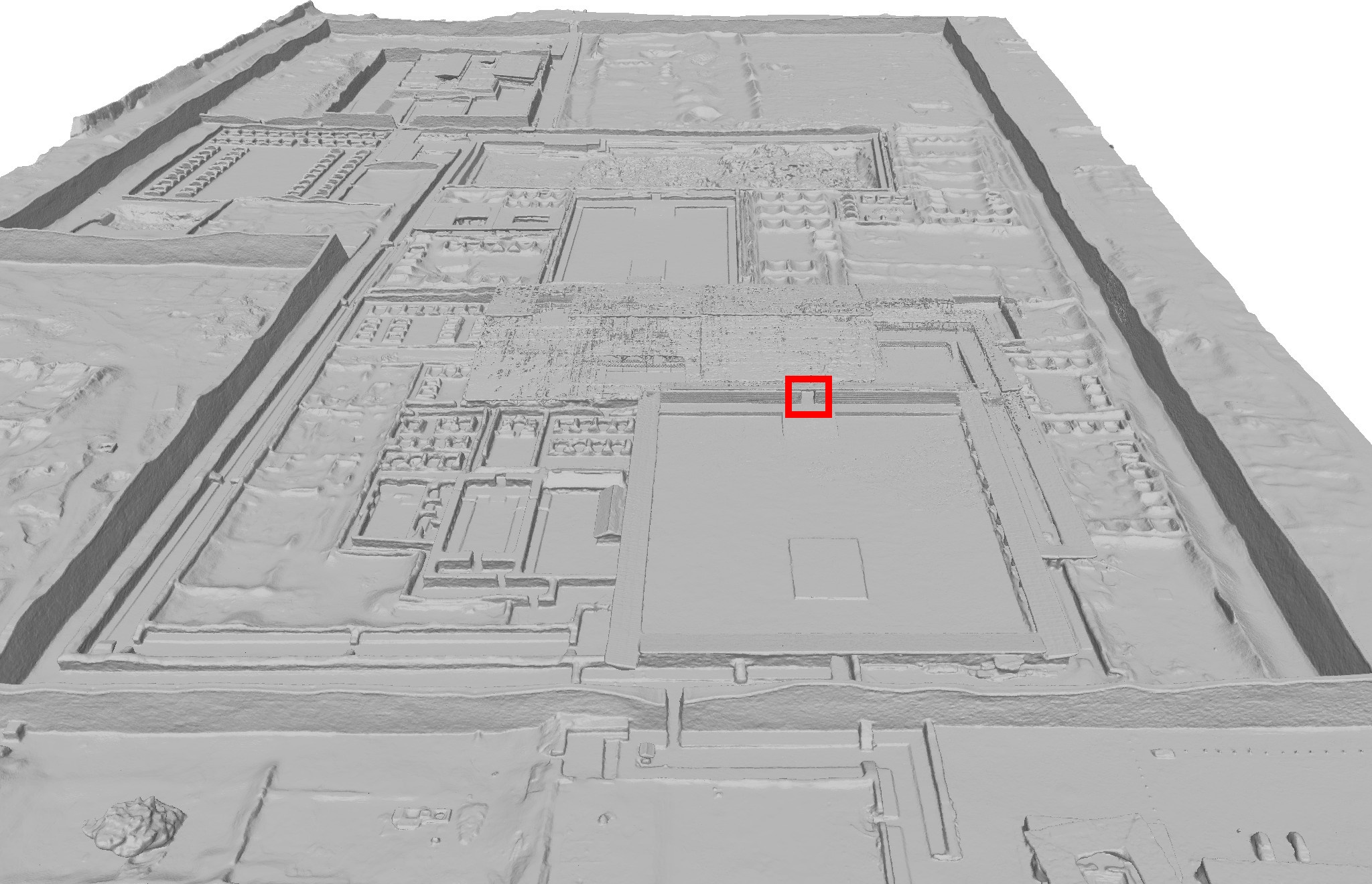}
    \end{minipage}
    \begin{minipage}[b]{0.49\linewidth}
        \includegraphics[width=\textwidth]{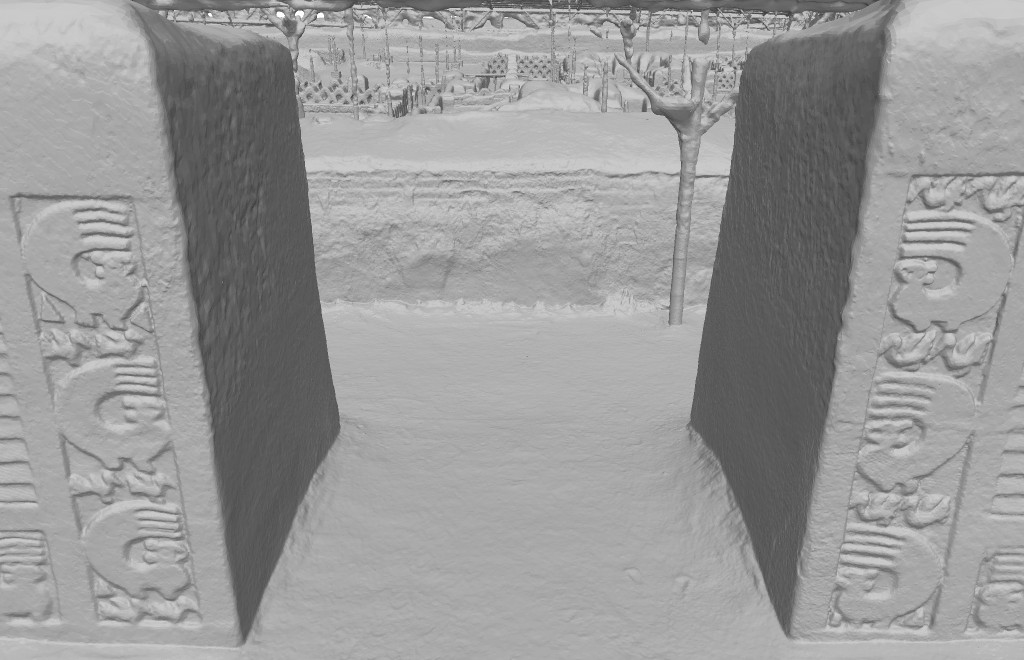}
    \end{minipage}
    \caption{Palacio Tschudi dataset and a closeup of an entrance.}
    \label{fig:palacio}
\end{figure}

\begin{table}

    \caption{
        Breakdown of Breisach dataset processing:
        2111 photos, 2642 million cubes from input depth maps, 4 hours 20 minutes of processing on a computer with an 8-core CPU and a GeForce GTX 1080 GPU with the peak RAM usage 10.07 GB.
    }

    \begin{tabular}{  |l||c|c| }
        \hline
        \textbf{Processing stage} & \textbf{Time} & \textbf{Time in \%} \\
        \hline
        Linear octree + merge & 30 + 11 min & 11\% + 4\% \\
        \hline
        Balance octree + merge & 7 + 11 min & 3\% + 4\% \\
        \hline
        Index treetop & 8 min & 3\%\\
        \hline
        Histograms (GPU) & 49 min & 19\%\\
        \hline
        Primal-dual method (GPU) & 88 min & 34\% \\
        \hline
        Marching cubes & 59 min & 22\% \\
        \hline
    \end{tabular}

    \label{tab:breisach_breakdown}

\end{table}

\section{Results}

We evaluated our method on an affordable computer with an 8-core CPU and a GeForce GTX 1080 GPU on five large datasets: Citywall\footnote{\url{https://www.gcc.tu-darmstadt.de/home/proj/mve/}} \cite{fuhrmann2014mve}
and Breisach\footnote{\url{https://lmb.informatik.uni-freiburg.de/people/ummenhof/multiscalefusion/}} \cite{ummenhofer2015global}
-- two datasets from previous papers with high scale diversity, Copenhagen\footnote{\url{https://download.kortforsyningen.dk/content/skraafoto}} \cite{skraafoto_copenhagen}
-- large-scale aerial photos of the city (this dataset was additionally evaluated on a small cluster too),
Palacio Tschudi\footnote{\url{https://doi.org/10.26301/4h29-7e80}} \cite{palacio_tschudi}
and Tomb of Tu Duc\footnote{\url{https://doi.org/10.26301/n06n-qa49}} \cite{tomb_of_tu_duc} (42 noise-free terrestrial LIDAR scans) -- two large public datasets collected by CyArk and distributed by Open Heritage 3D.

The summary for these datasets presented in Table~\ref{tab:datasets}.

For photo-based datasets, we executed the \textit{structure from motion} pipeline to estimate depth maps with SGM \cite{hirschmuller2007stereo}
method and evaluated our algorithm by using these depth maps as input. Note that to speed up the estimation of depthmaps, we downscaled original photos for some datasets -- see Table~\ref{tab:datasets}.
For the Tomb of Tu Duc LIDAR dataset, we converted each input scan into a 360-camera's depth map and used histogram votes with an increased weight -- see the listing in Algorithm \ref{alg:hists_listing}.
Processing breakdowns for other datasets including the Copenhagen city dataset (evaluated twice -- on an affordable computer and on a small cluster) with reconstruction results for many different city scenes are provided in the supplementary.

\begin{figure*}[hbt!]
    \centering
    \capstart
    \begin{minipage}[b]{0.23\linewidth}
        \begin{includegraphics}[width=\textwidth]{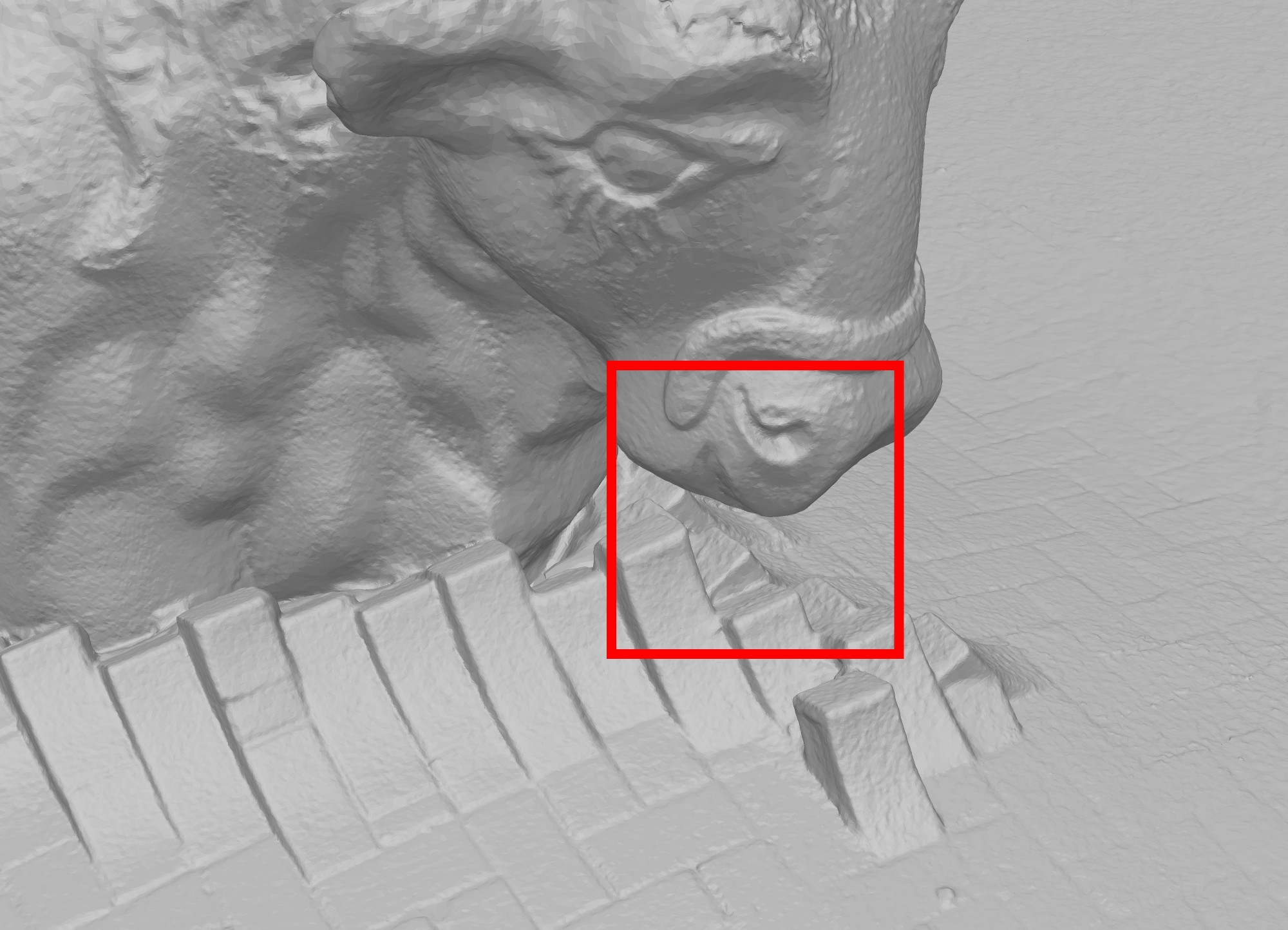}
            \put(-110,70){Ours}
        \end{includegraphics}
        \begin{includegraphics}[width=\textwidth]{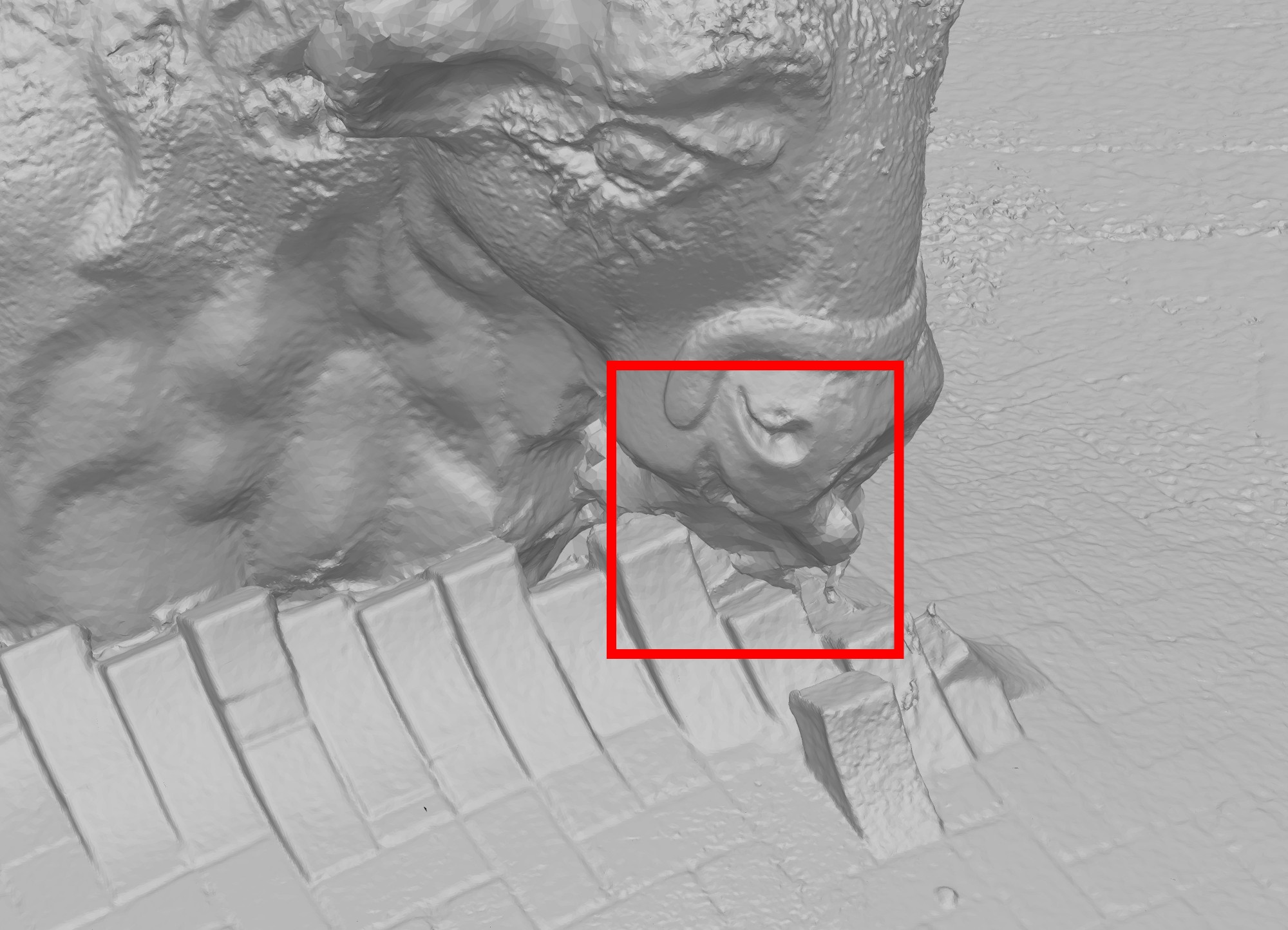}
            \put(-110,70){GDMR}
        \end{includegraphics}
    \end{minipage}
    \begin{minipage}[b]{0.51\linewidth}
        \includegraphics[width=\textwidth]{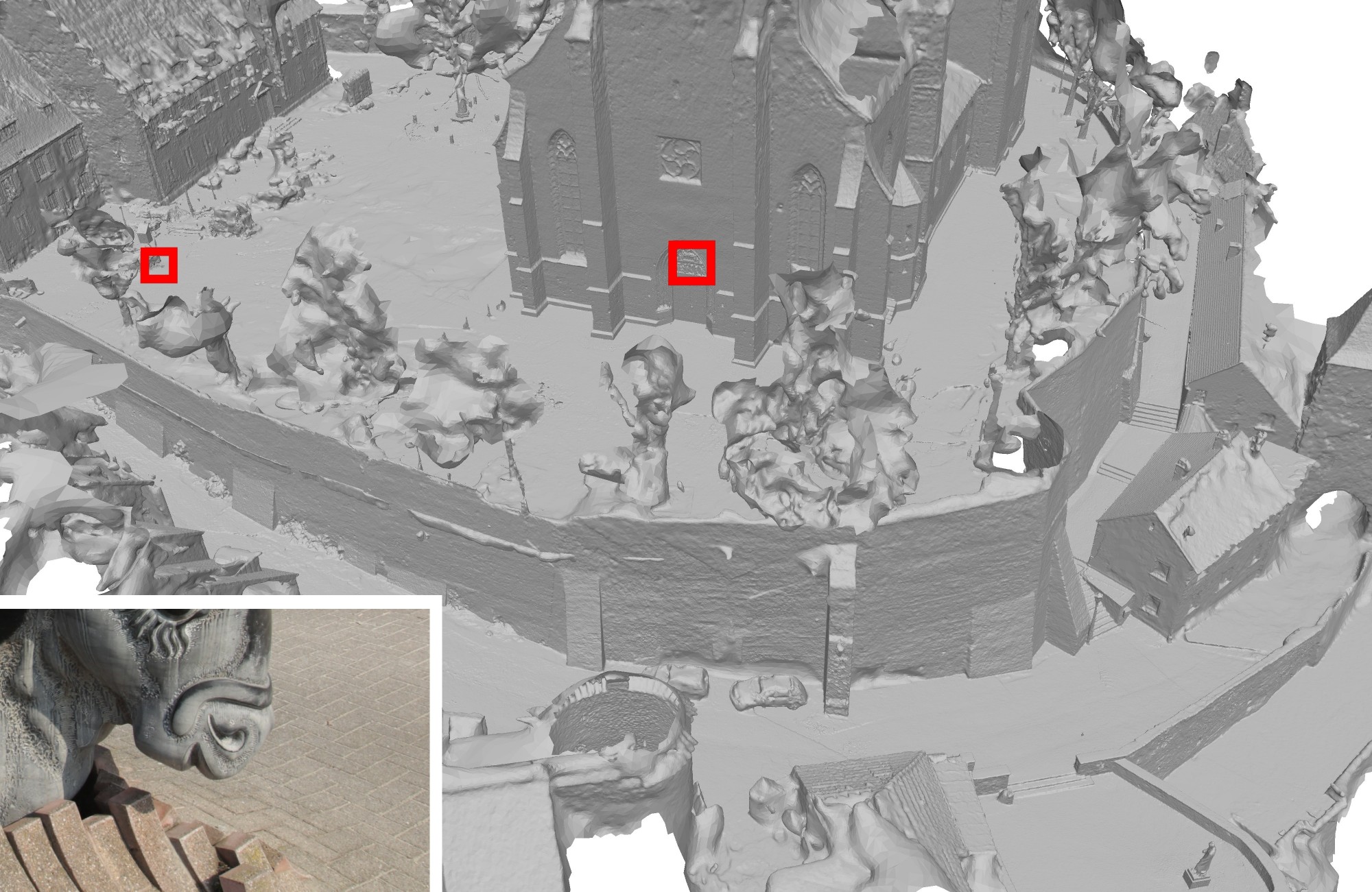}
    \end{minipage}
    \begin{minipage}[b]{0.23\linewidth}
        \begin{includegraphics}[width=\textwidth]{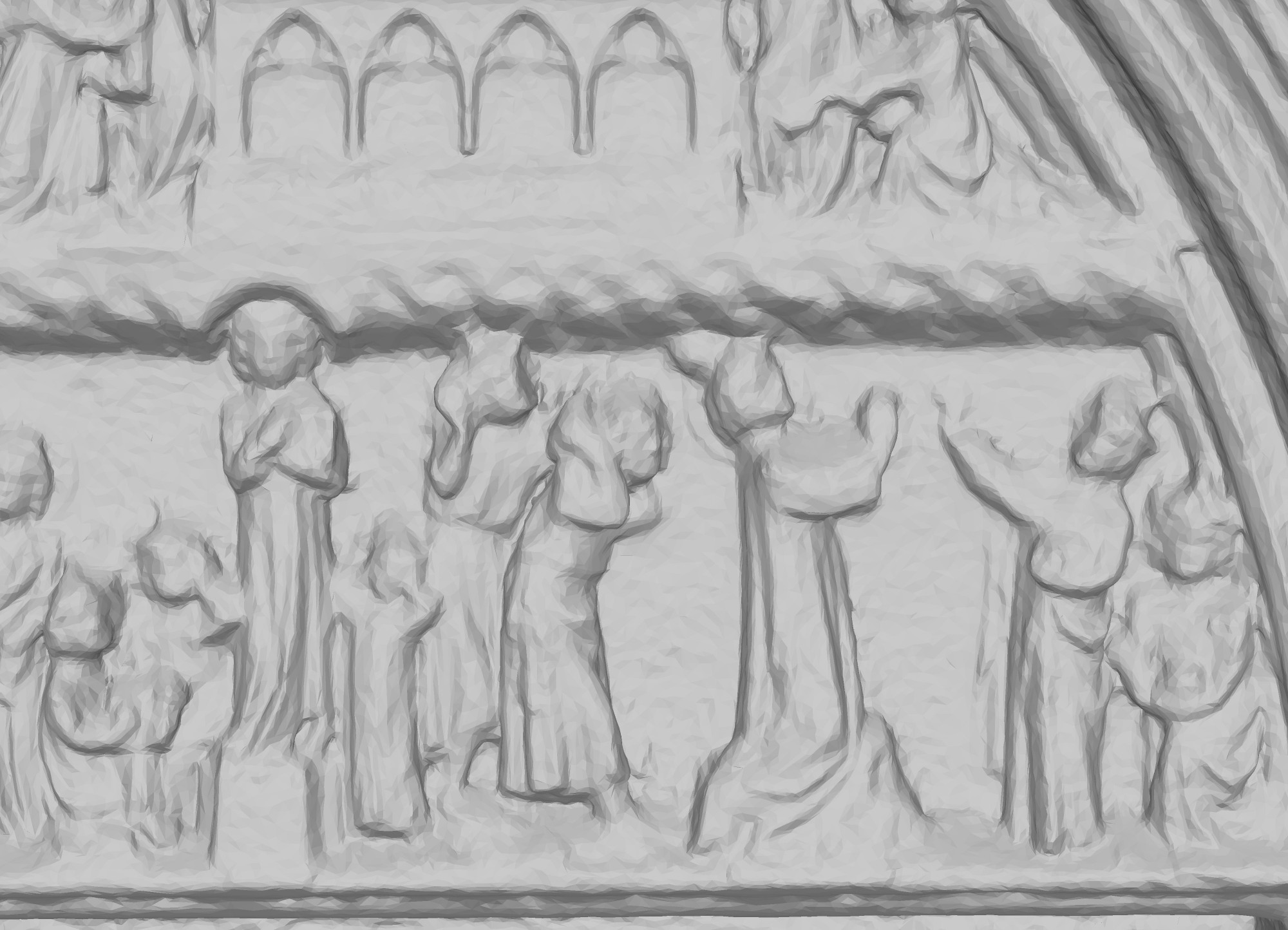}
            \put(-110,70){Ours}
        \end{includegraphics}
        \begin{includegraphics}[width=\textwidth]{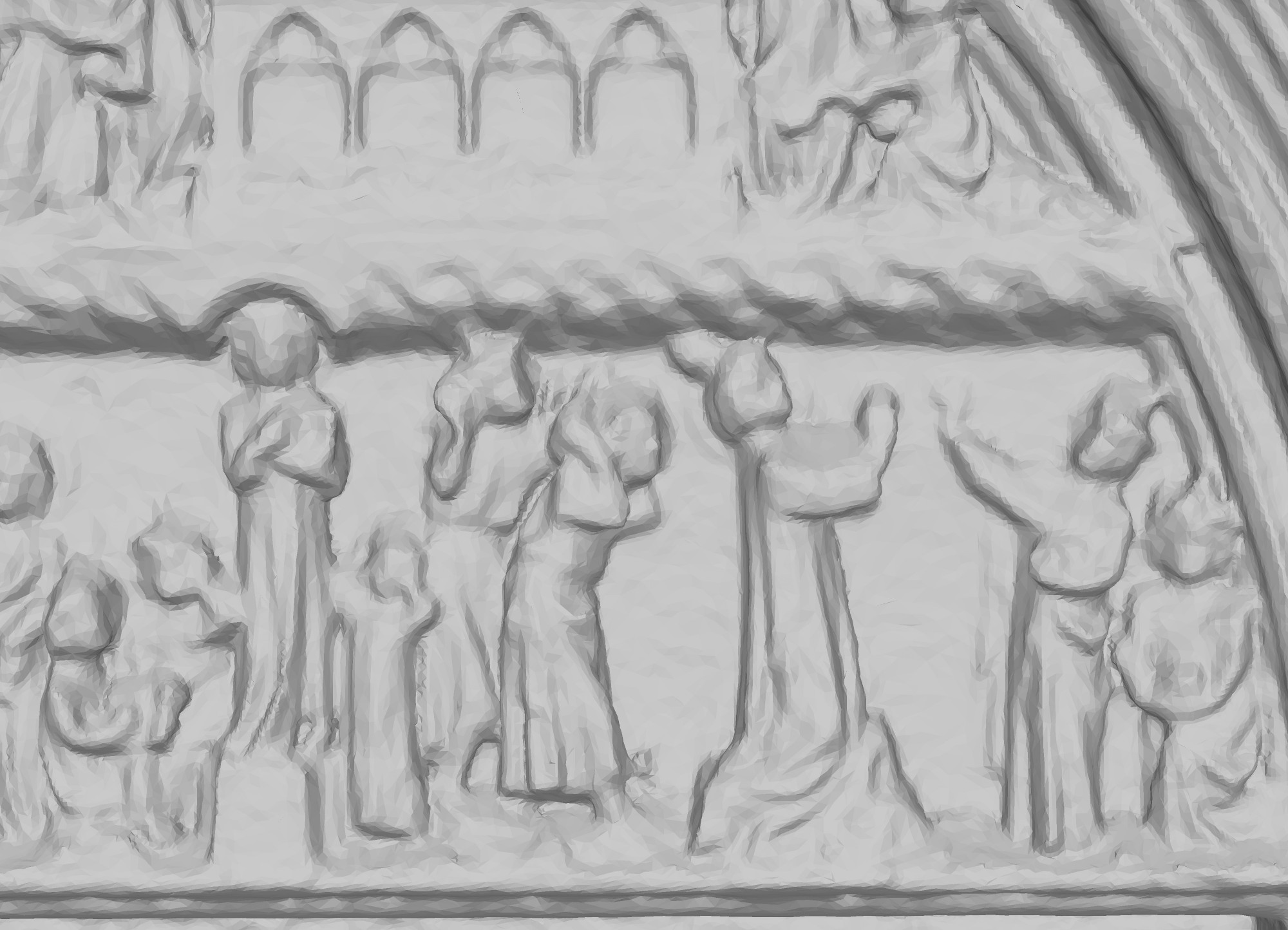}
            \put(-110,70){GDMR}
        \end{includegraphics}
    \end{minipage}
    \caption{Results on the Breisach dataset and comparison with GDMR \cite{ummenhofer2015global} results.
    The results have comparable quality.
    Note that thanks to strong visibility-based noise-filtering properties our method led to a cleaner space under the bull's nose.
        \vspace{0.05\textheight} 
    }
    \label{fig:breisach}
\end{figure*}

We ensured that the results are detailed and clean (see Fig.~\ref{fig:citywall},~\ref{fig:tomb_lidar},~\ref{fig:palacio},~\ref{fig:breisach}) for all datasets, and that our method's peak memory usage was between 10 GB and 17 GB.
Comparison with previous work \cite{ummenhofer2015global}, \cite{mostegel2017scalable}, presented in Table~\ref{tab:datasets_comparison}, shows that our method has significantly lower peak memory usage and is notably faster.
To ensure that this speedup was not at a cost of quality, we compared our results with  \cite{ummenhofer2015global} in Fig.~\ref{fig:citywall} and Fig.~\ref{fig:breisach}
(referred results were obtained with the software that their authors had used \footnote{We used pointfusion 0.2.0, publicly available at \url{http://lmb.informatik.uni-freiburg.de/people/ummenhof/multiscalefusion}}, we used the same depthmaps for quality comparison).

\begin{table*}[hbt!]

    \caption{List of presented datasets. For depthmaps estimation speedup, we downscaled original photos for some datasets before running SGM-based \cite{hirschmuller2007stereo} depthmap reconstruction.
    For example, original 2111 photos in Breisach dataset had the resolution of 5184x3456, and we downscaled them with x2 factor, down to 2592x1728 pixels.
    Note that the 'Initial cubes' column can be interpreted as 'the number of non-empty depth pixels in depthmaps', because each initial cube (before merging and octree balancing) corresponds to one sample from a depthmap.
    }

    \centering
    \begin{tabular}{  @{\extracolsep{\fill}} |l||c|c|c|c|c|c|c|c| @{} }
        \hline
        \thead{Dataset \\ name} & \thead{Images resolution \\ after downscale \\ (and downscale factor) } & \thead{Input \\ data} & \thead{Initial \\ cubes} & \thead{Merged and \\ balanced \\ cubes} & \thead{Faces after \\ marching cubes} & \thead{Decimated \\ faces} & \thead{Peak \\ RAM \\ (GB)} & \thead{Processing \\ time} \\
        \hline
        \shortstack{Citywall            \\  \cite{fuhrmann2014mve}} & \shortstack{2000x1500 \\ (x1)} & \shortstack{564 \\ depth maps} & 1205 mil & 404 mil & 135 mil & 15 mil & 13.17 & 63 min \\
        \hline
        \shortstack{Breisach            \\  \cite{ummenhofer2015global} } & \shortstack{2592x1728 \\ (x2)} & \shortstack{2111 \\ depth maps} & 2642 mil & 1457 mil & 558 mil & 57 mil & 10.07 & 260 min \\
        \hline
        \shortstack{Tomb of \\ Tu Duc   \\ (LIDAR) \\  \cite{tomb_of_tu_duc}} & \shortstack{8000x4000 \\ (x1)} & \shortstack{42 \\ LIDAR \\ scans} & 661 mil & 1304 mil & 672 mil & 48 mil & 10.05 & 160 min \\
        \hline
        \shortstack{Palacio \\ Tschudi  \\  \cite{palacio_tschudi}} & \shortstack{1840x1228 (37\%) \\ 1500x1000 (63\%) \\ (x4)} & \shortstack{13703 \\ depth maps} & 16 billion & 6 billion & 3159 mil & 243 mil & 16.75 & 1213 min \\
        \hline
        \shortstack{Copenha-\\gen city  \\  \cite{skraafoto_copenhagen}} & \shortstack{3368x2168 (26\%) \\ 2575x1925 (74\%) \\ (x4)} & \shortstack{27472 \\depth maps} & 28 billion & 24 billion & 7490 mil & 267 mil & 13.35 & 1758 min \\
        \hline
    \end{tabular}

    \label{tab:datasets}

\end{table*}

\begin{table*}[hbt!]

    \caption{Comparison with the previous results - GDMR \cite{ummenhofer2015global} and SSR 128K \cite{mostegel2017scalable}. Note that SSR had 8.9 GB per-thread peak memory, and finished the reconstruction in 58.3 hours using 32 threads, so total peak memory could be estimated as about 32*8.9=285 GB.
    Also note that GDMR results for Breisach dataset were taken from the original paper, but because the authors did not mention memory and time results for the Citywall dataset -- we provide our results of GDMR evaluation on a computer with an 8-core CPU starting from 1205 million input points.}

    \centering
    \begin{tabular}{  @{\extracolsep{\fill}} |l||c|c|c|c|c|c|c| @{} }
        \hline
        \thead{Dataset \\ name} & \thead{Input \\ data} & \thead{GDMR \\ Peak RAM} & \thead{GDMR \\ time} & \thead{\textbf{Our} \\ Peak RAM} & \thead{\textbf{Our} \\ time} & \thead{SSR \\ Peak RAM} & \thead{SSR \\ time} \\
        \hline
        \shortstack{Citywall} & \shortstack{564 \\ depth maps} & 75 GB & 19 h & 13.17 GB & 63 min & 32*8.9 GB & 58 h \\
        \hline
        \shortstack{Breisach} & \shortstack{2111 \\ depth maps} & 64 GB & 76 h & 10.07 GB & 260 min & N/A & N/A \\
        \hline
    \end{tabular}

    \label{tab:datasets_comparison}

\end{table*}

\section{Conclusions}

In this work, we present an out-of-core method for surface reconstruction from depth maps and terrestrial LIDAR scans.
Our results have shown that the algorithm specifics do not increase the running time; instead,
thanks to GPU-acceleration our implementation has proven to be much faster than previously published results on the datasets that we have used for testing.
We have also shown that the quality of results is comparable to an in-core reconstruction method GDMR \cite{ummenhofer2015global}.
Note that an out-of-core balanced octree with treetop indexing
is a rather general concept that can be used as a framework for different methods in a similar way as we used it with the $TGV$ minimization method.

One of the main contributions of our work is an out-of-core framework for fast and detailed surface reconstruction.
\ificcvfinal Our method is available as part of commercial software.\fi

\vspace{0.2\textheight} 
\bibliographystyle{ieee_fullname}
\bibliography{egbib}

\begin{thebibliography}{10}\itemsep=-1pt

\bibitem{bloomenthal1988polygonization}
Jules Bloomenthal.
\newblock {\em Polygonization of implicit surfaces}.
\newblock Citeseer, 1988.

\bibitem{bolitho2007multilevel}
Matthew Bolitho, Michael Kazhdan, Randal Burns, and Hugues Hoppe.
\newblock Multilevel streaming for out-of-core surface reconstruction.
\newblock pages 69--78, 2007.

\bibitem{boykov2004experimental}
Yuri Boykov and Vladimir Kolmogorov.
\newblock An experimental comparison of min-cut/max-flow algorithms for energy
  minimization in vision.
\newblock {\em IEEE transactions on pattern analysis and machine intelligence},
  26(9):1124--1137, 2004.

\bibitem{bredies2010total}
Kristian Bredies, Karl Kunisch, and Thomas Pock.
\newblock Total generalized variation.
\newblock {\em SIAM Journal on Imaging Sciences}, 3(3):492--526, 2010.

\bibitem{curless1996volumetric}
Brian Curless and Marc Levoy.
\newblock A volumetric method for building complex models from range images.
\newblock In {\em Proceedings of the 23rd annual conference on Computer
  graphics and interactive techniques}, pages 303--312, 1996.

\bibitem{tomb_of_tu_duc}
CyArk.
\newblock Complex of hué monuments - tomb of tu duc, vietnam, 2019.

\bibitem{palacio_tschudi}
CyArk.
\newblock Palacio tschudi - chan chan, peru, 2020.

\bibitem{skraafoto_copenhagen}
{Danish Agency for Data Supply and Efficiency}.
\newblock Skraafoto of copenhagen, 2019.

\bibitem{fuhrmann2014floating}
Simon Fuhrmann and Michael Goesele.
\newblock Floating scale surface reconstruction.
\newblock {\em ACM Transactions on Graphics (ToG)}, 33(4):1--11, 2014.

\bibitem{fuhrmann2014mve}
Simon Fuhrmann, Fabian Langguth, and Michael Goesele.
\newblock Mve-a multi-view reconstruction environment.
\newblock In {\em GCH}, pages 11--18, 2014.

\bibitem{garland1998simplifying}
Michael Garland and Paul~S Heckbert.
\newblock Simplifying surfaces with color and texture using quadric error
  metrics.
\newblock In {\em Proceedings Visualization'98 (Cat. No. 98CB36276)}, pages
  263--269. IEEE, 1998.

\bibitem{goldberg2015faster}
Andrew~V Goldberg, Sagi Hed, Haim Kaplan, Pushmeet Kohli, Robert~E Tarjan, and
  Renato~F Werneck.
\newblock Faster and more dynamic maximum flow by incremental breadth-first
  search.
\newblock In {\em Algorithms-ESA 2015}, pages 619--630. Springer, 2015.

\bibitem{graber2011online}
Gottfried Graber, Thomas Pock, and Horst Bischof.
\newblock Online 3d reconstruction using convex optimization.
\newblock pages 708--711, 2011.

\bibitem{han2019scalable}
Jiali Han and Shuhan Shen.
\newblock Scalable point cloud meshing for image-based large-scale 3d modeling.
\newblock {\em Visual Computing for Industry, Biomedicine, and Art}, 2(1):1--9,
  2019.

\bibitem{hiep2009towards}
Vu~Hoang Hiep, Renaud Keriven, Patrick Labatut, and Jean-Philippe Pons.
\newblock Towards high-resolution large-scale multi-view stereo.
\newblock pages 1430--1437, 2009.

\bibitem{hirschmuller2007stereo}
Heiko Hirschmuller.
\newblock Stereo processing by semiglobal matching and mutual information.
\newblock {\em IEEE Transactions on pattern analysis and machine intelligence},
  30(2):328--341, 2007.

\bibitem{jancosek2014exploiting}
Michal Jancosek and Tomas Pajdla.
\newblock Exploiting visibility information in surface reconstruction to
  preserve weakly supported surfaces.
\newblock {\em International scholarly research notices}, 2014, 2014.

\bibitem{kazhdan2006poisson}
Michael Kazhdan, Matthew Bolitho, and Hugues Hoppe.
\newblock Poisson surface reconstruction.
\newblock 7, 2006.

\bibitem{kuhn2017tv}
Andreas Kuhn, Heiko Hirschm{\"u}ller, Daniel Scharstein, and Helmut Mayer.
\newblock A tv prior for high-quality scalable multi-view stereo
  reconstruction.
\newblock {\em International Journal of Computer Vision}, 124(1):2--17, 2017.

\bibitem{kuhn2015incremental}
Andreas Kuhn and Helmut Mayer.
\newblock Incremental division of very large point clouds for scalable 3d
  surface reconstruction.
\newblock pages 10--18, 2015.

\bibitem{li2016efficient}
Shiwei Li, Sing~Yu Siu, Tian Fang, and Long Quan.
\newblock Efficient multi-view surface refinement with adaptive resolution
  control.
\newblock pages 349--364, 2016.

\bibitem{merrell2007real}
Paul Merrell, Amir Akbarzadeh, Liang Wang, Philippos Mordohai, Jan-Michael
  Frahm, Ruigang Yang, David Nist{\'e}r, and Marc Pollefeys.
\newblock Real-time visibility-based fusion of depth maps.
\newblock pages 1--8, 2007.

\bibitem{morton1966computer}
Guy~M Morton.
\newblock A computer oriented geodetic data base and a new technique in file
  sequencing.
\newblock 1966.

\bibitem{mostegel2017scalable}
Christian Mostegel, Rudolf Prettenthaler, Friedrich Fraundorfer, and Horst
  Bischof.
\newblock Scalable surface reconstruction from point clouds with extreme scale
  and density diversity.
\newblock pages 904--913, 2017.

\bibitem{pock2011tgv}
Thomas Pock, Lukas Zebedin, and Horst Bischof.
\newblock Tgv-fusion.
\newblock pages 245--258, 2011.

\bibitem{tu2004balance}
Tiankai Tu and David~R O’hallaron.
\newblock Balance refinement of massive linear octree.
\newblock 2004.

\bibitem{ummenhofer2015global}
Benjamin Ummenhofer and Thomas Brox.
\newblock Global, dense multiscale reconstruction for a billion points.
\newblock pages 1341--1349, 2015.

\bibitem{ummenhofer2017global}
Benjamin Ummenhofer and Thomas Brox.
\newblock Global, dense multiscale reconstruction for a billion points.
\newblock {\em International Journal of Computer Vision}, pages 1--13, 2017.

\bibitem{vu2011high}
Hoang-Hiep Vu, Patrick Labatut, Jean-Philippe Pons, and Renaud Keriven.
\newblock High accuracy and visibility-consistent dense multiview stereo.
\newblock {\em IEEE transactions on pattern analysis and machine intelligence},
  34(5):889--901, 2011.

\bibitem{williams1983pyramidal}
Lance Williams.
\newblock Pyramidal parametrics.
\newblock In {\em Proceedings of the 10th annual conference on Computer
  graphics and interactive techniques}, pages 1--11, 1983.

\bibitem{zach2008fast}
Christopher Zach.
\newblock Fast and high quality fusion of depth maps.
\newblock 1(2), 2008.

\bibitem{zach2007globally}
Christopher Zach, Thomas Pock, and Horst Bischof.
\newblock A globally optimal algorithm for robust tv-l1 range image
  integration.
\newblock pages 1--8, 2007.

\bibitem{zhou2019detail}
Yang Zhou, Shuhan Shen, and Zhanyi Hu.
\newblock Detail preserved surface reconstruction from point cloud.
\newblock {\em Sensors}, 19(6):1278, 2019.

\end{thebibliography}


\begin{thebibliography}{1}\itemsep=-1pt

\bibitem{skraafoto_copenhagen}
{Danish Agency for Data Supply and Efficiency}.
\newblock Skraafoto of copenhagen, 2019.

\bibitem{hirschmuller2007stereo}
Heiko Hirschmuller.
\newblock Stereo processing by semiglobal matching and mutual information.
\newblock {\em IEEE Transactions on pattern analysis and machine intelligence},
  30(2):328--341, 2007.

\bibitem{gmaps_copenhagen}
Google Maps.
\newblock {Copenhagen, Denmark, 55°40'33.95''N 12°34'6.01''E}, 2020.

\end{thebibliography}

\end{document}


\title{Building Copenhagen in a Day\footnote{In 29 hours}.
\\ Supplementary Material for: \\ Out-of-Core Surface Reconstruction via Global $TGV$ Minimization.} 

\author{Nikolai Poliarnyi\\
    Agisoft LLC, St. Petersburg, Russia\\
    {\tt\small polarnick@agisoft.com}
}

\maketitle

In this supplementary, we want to show a public dataset with aerial photos of Copenhagen that we processed with our out-of-core surface reconstruction method.
Additionally, we provide closeups of the resulting model with per-vertex colors and breakdowns of other three datasets discussed in the paper -- see Tables~\ref{tab:citywall_breakdown}, ~\ref{tab:palacio_breakdown}, ~\ref{tab:tomb_breakdown}.

\begin{figure}[H]
    \centering
    \capstart
    \begin{minipage}[b]{0.45\linewidth}
        \includegraphics[width=\textwidth]{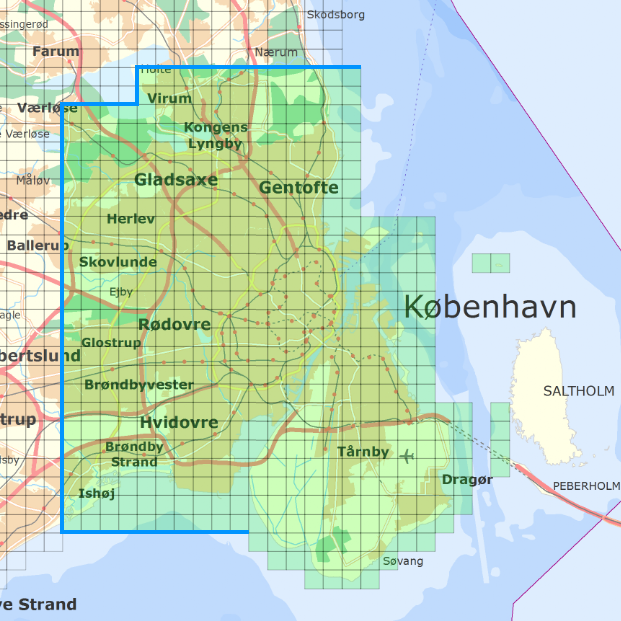}
    \end{minipage}
    \begin{minipage}[b]{0.45\linewidth}
        \includegraphics[width=\textwidth]{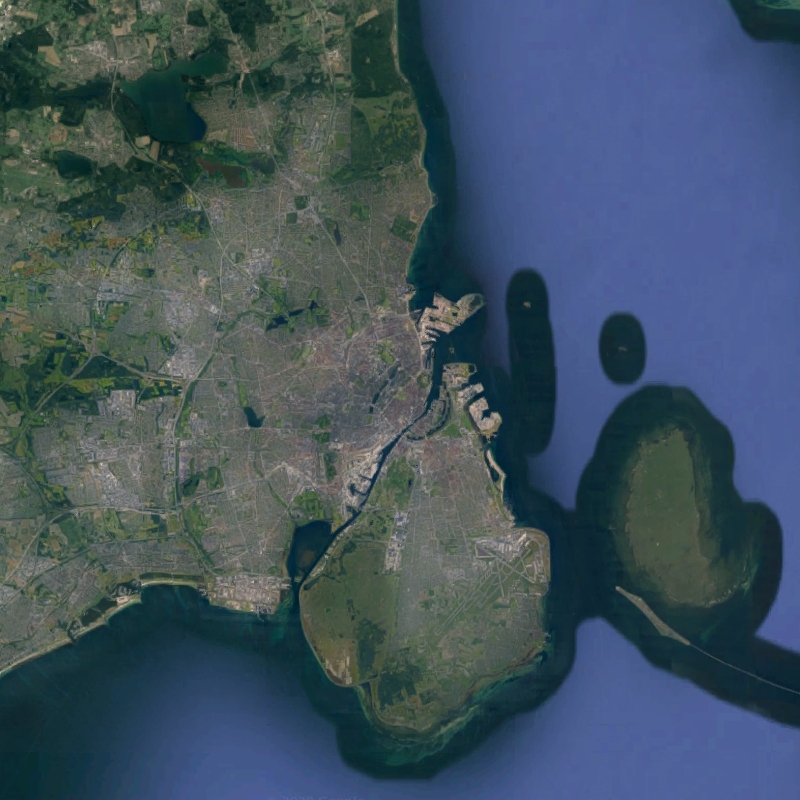}
    \end{minipage}
    \caption{Copenhagen dataset \cite{skraafoto_copenhagen}: 500 processed green blocks are shown (outlined with a blue border). To the right, a corresonding google map is shown \cite{gmaps_copenhagen}.}
    \label{fig:copenhagen_tiles}
\end{figure}

\begin{figure}
    \centering
    \capstart
    \begin{minipage}[b]{0.32\linewidth}
        \includegraphics[width=\textwidth]{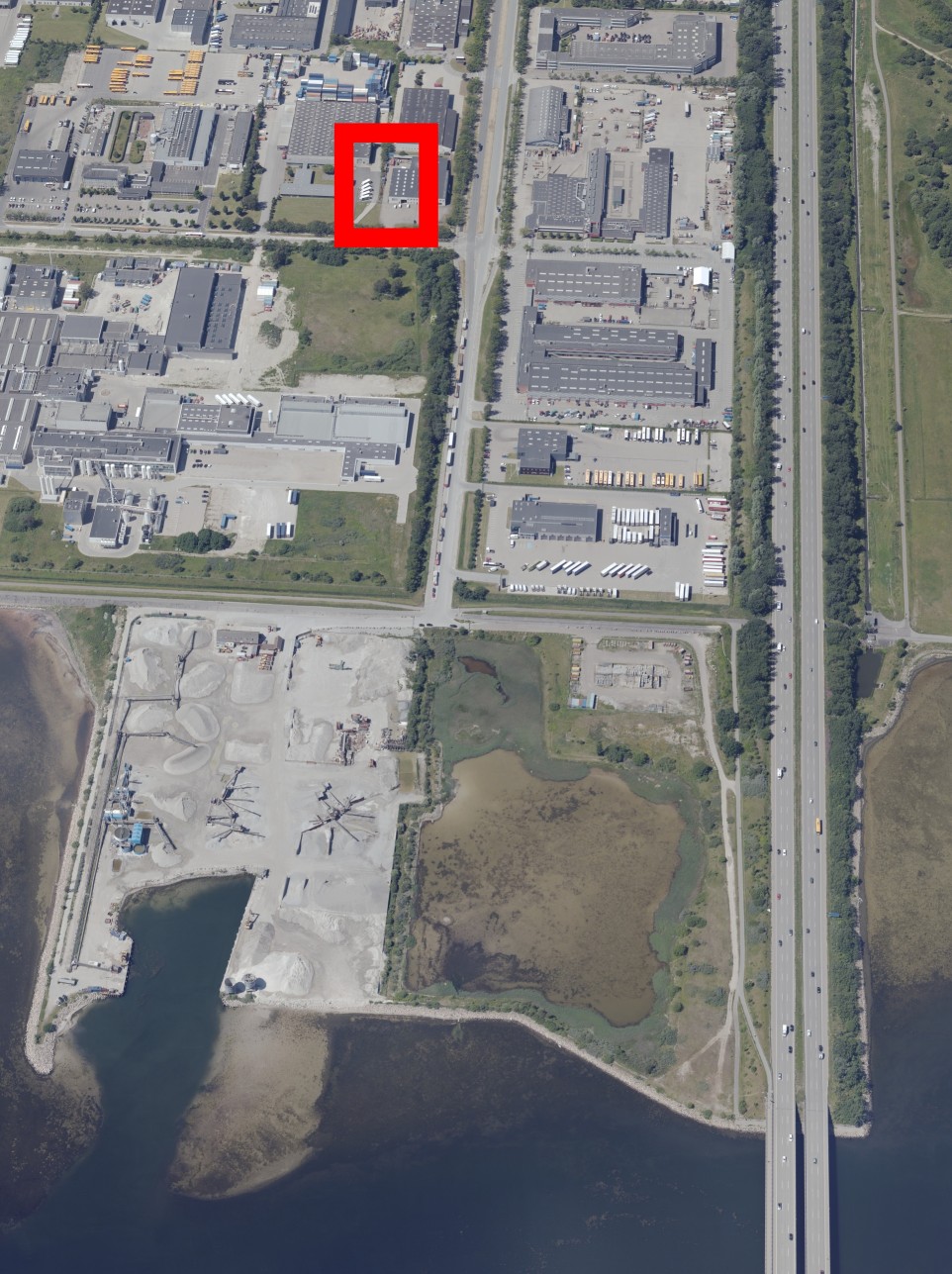}
    \end{minipage}
    \begin{minipage}[b]{0.32\linewidth}
        \includegraphics[width=\textwidth]{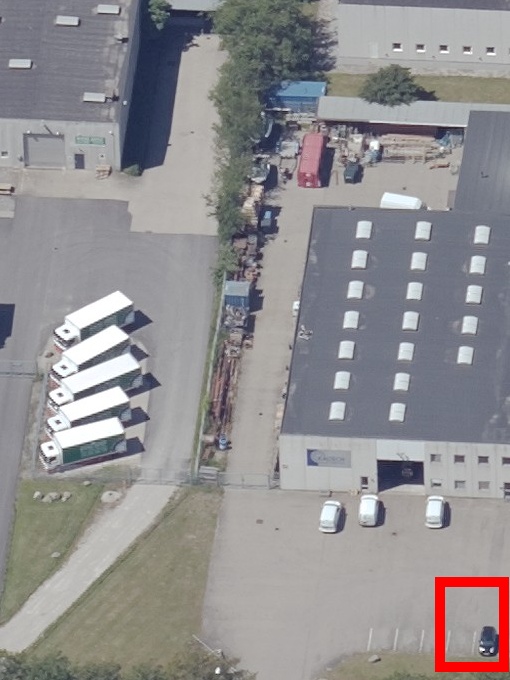}
    \end{minipage}
    \begin{minipage}[b]{0.32\linewidth}
        \includegraphics[width=\textwidth]{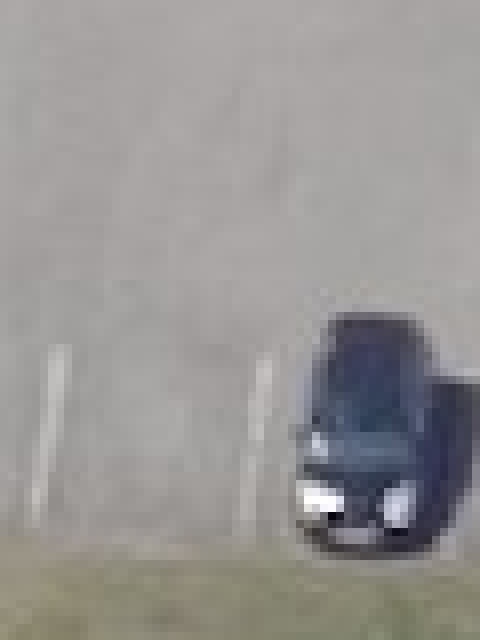}
    \end{minipage}
    \caption{Example of oblique photo with resolution 7700x10300. skraa\_1km\_6168\_720\_JPG\_UTM32-ETRS89.zip / 2019\_84\_40\_5\_0034\_00002391.jpg}
    \label{fig:oblique_photo}
\end{figure}

We downloaded 500 blocks (tiles) of photos covering \SI{425}{\km\squared} of the Copenhagen city from publicly available aerial photos of Denmark\footnote{\url{https://download.kortforsyningen.dk/content/skraafoto}} \cite{skraafoto_copenhagen}.
The downloaded blocks are shown in Fig.~\ref{fig:copenhagen_tiles}. For convenience, their names are also listed in the attached supplementary text file blocks.txt.
These 500 blocks (each covers \SI{1}{\km\squared}) contain 27472 aerial photos (566 GB) with resolutions of 13470x8670 and 10300x7700.
In addition to nadir photos, there is also oblique imagery -- one of such photos is shown in Fig.~\ref{fig:oblique_photo}.

\begin{figure}[H]
    \centering
    \capstart
    \begin{minipage}[b]{1.0\linewidth}
        \includegraphics[width=\textwidth]{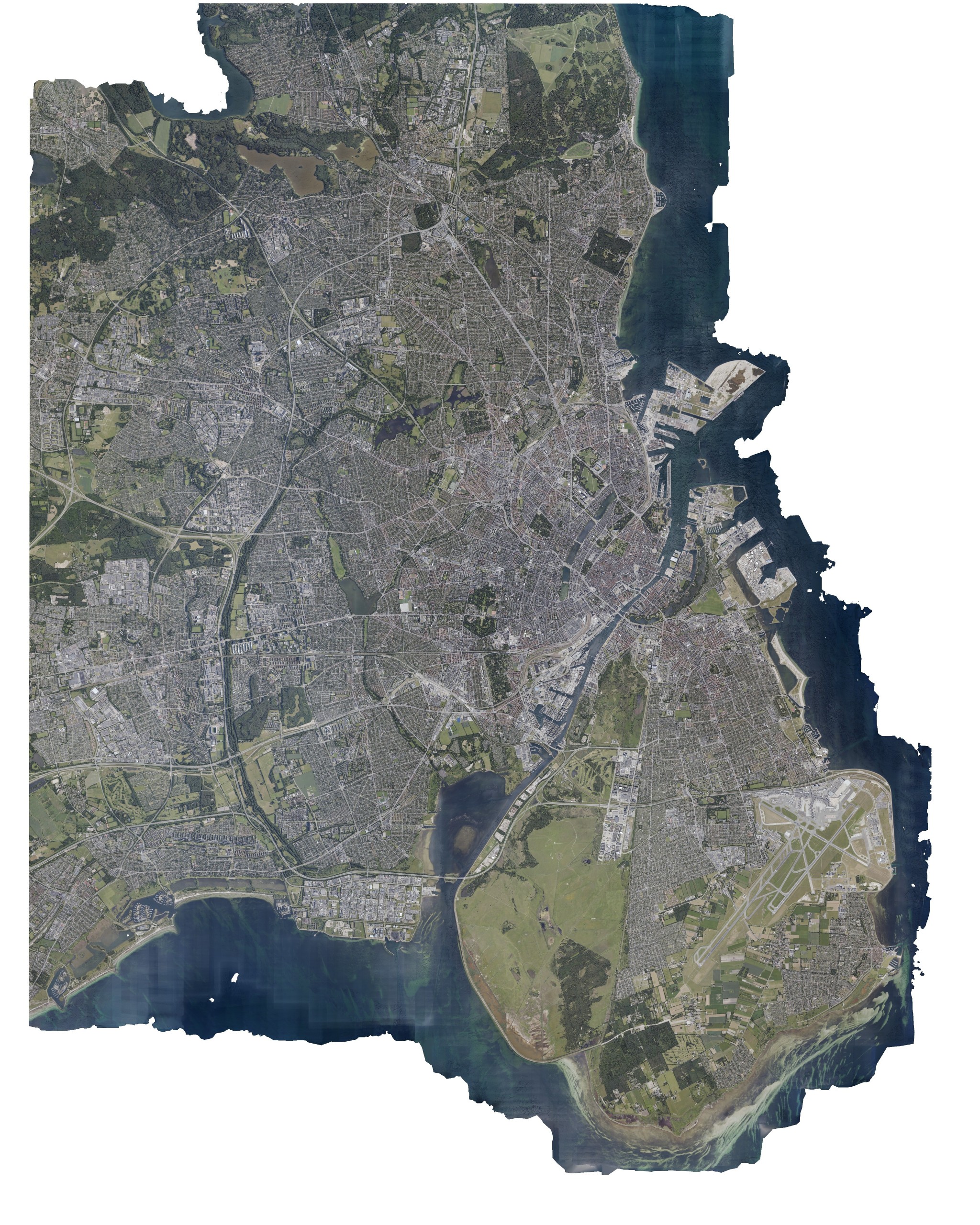}
    \end{minipage}
    \caption{Our method can handle an arbitrary large scene -- even \SI{425}{\km\squared} of the Copenhagen city.
    This polygonal model, consisting of 267 million triangles with per-vertex colors, was reconstructed from depth maps of 27472 aerial photos in 29 hours.}
    \label{fig:copenhagen_full_rgb}
\end{figure}

For speedup, we downscaled these photos with a factor of 4 before running the SGM-based \cite{hirschmuller2007stereo} depthmap reconstruction.
This dataset was evaluated twice - on a small cluster of 7 affordable computers (on the average each computer was a PC with an 8-core CPU and a GeForce GTX 1080 GPU)
and on a single computer with an 8-core CPU and a GeForce GTX 1080 GPU.
The surface reconstruction took 29 hours on a small cluster of 7 computers, and the processing took 5.5 days on a single computer.
The breakdown of the processing on a cluster is shown in Table~\ref{tab:copenhagen_breakdown}.
The whole model with per-vertex color is shown in Fig.~\ref{fig:copenhagen_full_rgb}.

\begin{table*}

    \caption{
        A breakdown of the Copenhagen city dataset processing:
        27472 photos, 28 billion cubes from input depth maps, 29 hours of processing on a 7-computers cluster with the peak RAM usage -- 13.35 GB.
    }

    \centering
    \begin{tabular*}{0.55\columnwidth}{  @{\extracolsep{\fill}} |l||c|c| @{} }
        \hline
        \textbf{Processing stage} & \textbf{Time} & \textbf{Time in \%} \\
        \hline
        Linear octree parts + merge & 173 + 149 min & 10\% + 8\% \\
        \hline
        Balance octree parts + merge & 141 + 199 min & 8\% + 11\% \\
        \hline
        Index treetop & 80 min & 5\%\\
        \hline
        Histograms (GPU) & 393 min & 22\%\\
        \hline
        Primal-dual method (GPU) & 159 min & 9\% \\
        \hline
        Marching cubes & 445 min & 25\% \\
        \hline
    \end{tabular*}

    \label{tab:copenhagen_breakdown}

\end{table*}

\begin{table}

    \caption{
        A breakdown of the Citywall dataset processing:
        564 photos, 1205 million cubes from input depth maps, 63 minutes of processing on a computer with a 8-core CPU and a GeForce GTX 1080 GPU with the peak RAM usage -- 13.17 GB.
    }

    \centering
    \begin{tabular*}{0.55\columnwidth}{  @{\extracolsep{\fill}} |l||c|c| @{} }
        \hline
        \textbf{Processing stage} & \textbf{Time} & \textbf{Time in \%} \\
        \hline
        Linear octree + merge & 7 + 4 min & 10\% + 6\%\\
        \hline
        Balance octree + merge & 2 + 2 min & 3\% + 3\%\\
        \hline
        Index treetop & 2 min & 3\%\\
        \hline
        Histograms (GPU) & 17 min & 26\%\\
        \hline
        Primal-dual method (GPU) & 18 min & 28\% \\
        \hline
        Marching cubes & 12 min & 18\% \\
        \hline
    \end{tabular*}

    \label{tab:citywall_breakdown}

\end{table}

\begin{table}

    \caption{
        A breakdown of the Palacio Tschudi dataset processing:
        13703 photos, 16 billion cubes from input depth maps, 20 hours of processing on a computer with an 8-core CPU and a GeForce GTX 1080 GPU with the peak RAM usage -- 16.75 GB.
    }

    \centering
    \begin{tabular*}{0.55\columnwidth}{  @{\extracolsep{\fill}} |l||c|c| @{} }
        \hline
        \textbf{Processing stage} & \textbf{Time} & \textbf{Time in \%} \\
        \hline
        Linear octree + merge & 66 + 68 min & 5\% + 5\%\\
        \hline
        Balance octree + merge & 28 + 56 min & 2\% + 5\%\\
        \hline
        Index treetop & 32 min & 2\%\\
        \hline
        Histograms (GPU) & 370 min & 30\%\\
        \hline
        Primal-dual method (GPU) & 360 min & 30\% \\
        \hline
        Marching cubes & 240 min & 20\% \\
        \hline
    \end{tabular*}

    \label{tab:palacio_breakdown}

\end{table}

\begin{table}

    \caption{
        A breakdown of the Tomb of Tu Duc LIDAR dataset processing:
        42 LIDAR scans, 661 million cubes from input LIDAR scans, 160 minutes of processing on a computer with an 8-core CPU with a GeForce GTX 1080 GPU with peak RAM usage -- 10.05 GB.
    }

    \centering
    \begin{tabular*}{0.55\columnwidth}{  @{\extracolsep{\fill}} |l||c|c| @{} }
        \hline
        \textbf{Processing stage} & \textbf{Time} & \textbf{Time in \%} \\
        \hline
        Linear octree + merge & 4 + 3 min & 3\% + 2\%\\
        \hline
        Balance octree + merge & 5 + 21 min & 3\% + 13\%\\
        \hline
        Index treetop & 6 min & 4\%\\
        \hline
        Histograms (GPU) & 14 min & 9\%\\
        \hline
        Primal-dual method (GPU) & 54 min & 34\% \\
        \hline
        Marching cubes & 53 min & 33\% \\
        \hline
    \end{tabular*}

    \label{tab:tomb_breakdown}

\end{table}

\begin{figure}
    \centering
    \capstart
    \begin{minipage}[b]{1.0\linewidth}
        \includegraphics[width=\textwidth]{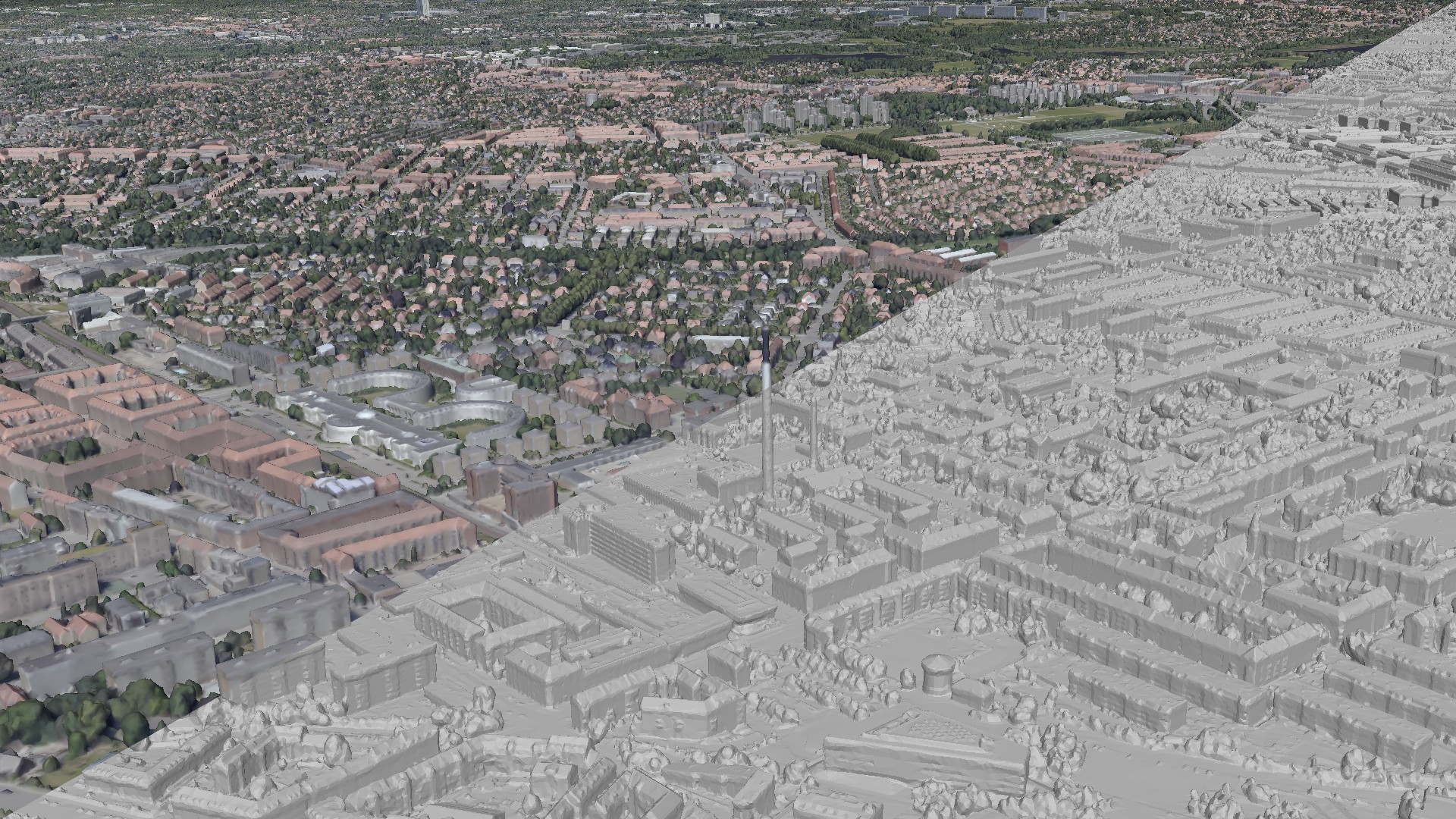}
    \end{minipage}
    \caption{Frederiksberg Forsyning A/S closeup. Note that both pipes were reconstructed well.}
    \label{fig:closeup_frederiksberg}
\end{figure}

\begin{figure}
    \centering
    \capstart
    \begin{minipage}[b]{1.0\linewidth}
        \includegraphics[width=\textwidth]{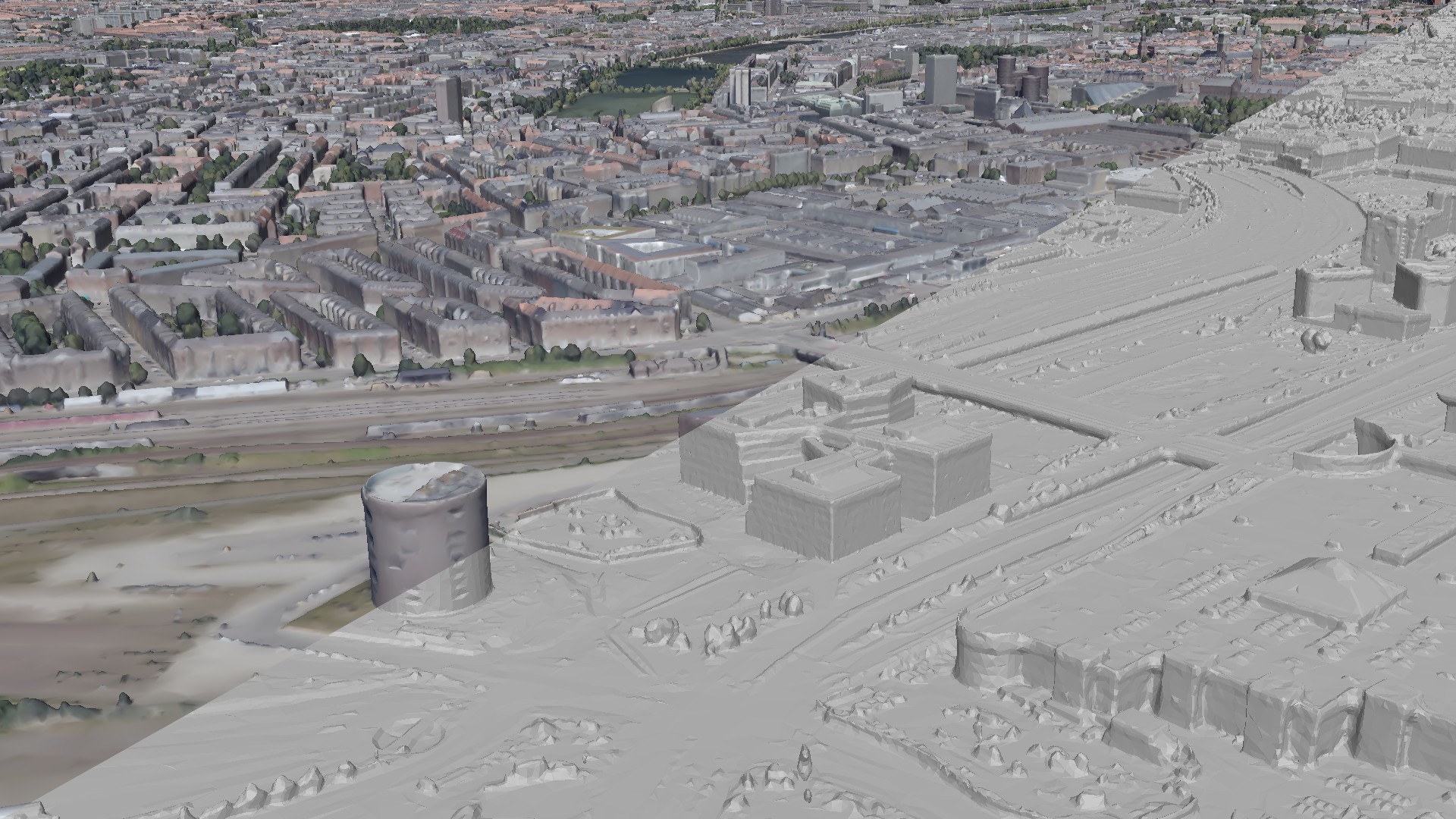}
    \end{minipage}
    \caption{Banedanmark closeup.}
    \label{fig:closeup_banedanmark}
\end{figure}

\begin{figure}
    \centering
    \capstart
    \begin{minipage}[b]{1.0\linewidth}
        \includegraphics[width=\textwidth]{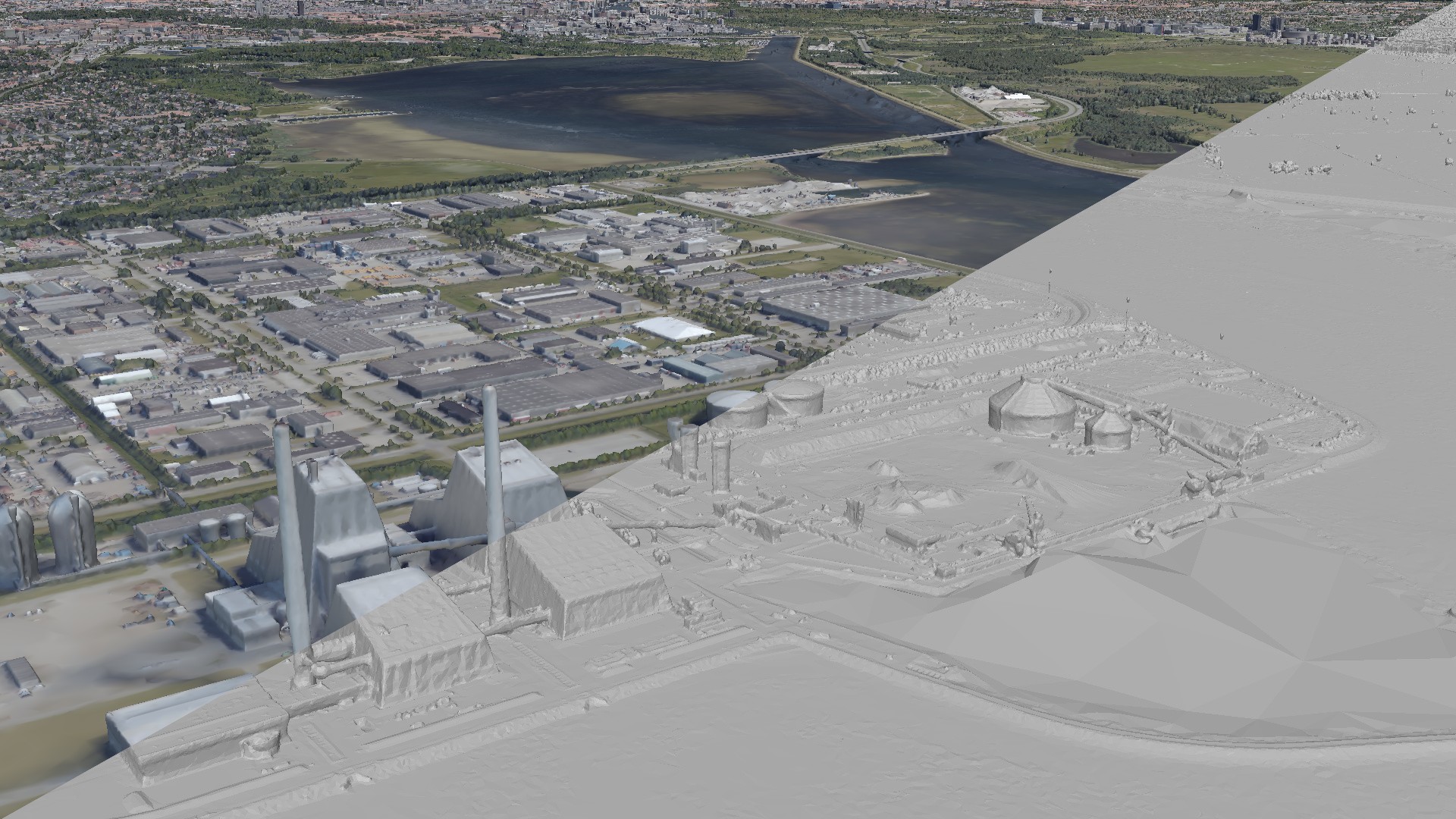}
    \end{minipage}
    \caption{Aved{\o}re Power Station closeup. Note accurate geometry of thin structures above the ground.}
    \label{fig:closeup_power_station}
\end{figure}

\begin{figure}
    \centering
    \capstart
    \begin{minipage}[b]{1.0\linewidth}
        \includegraphics[width=\textwidth]{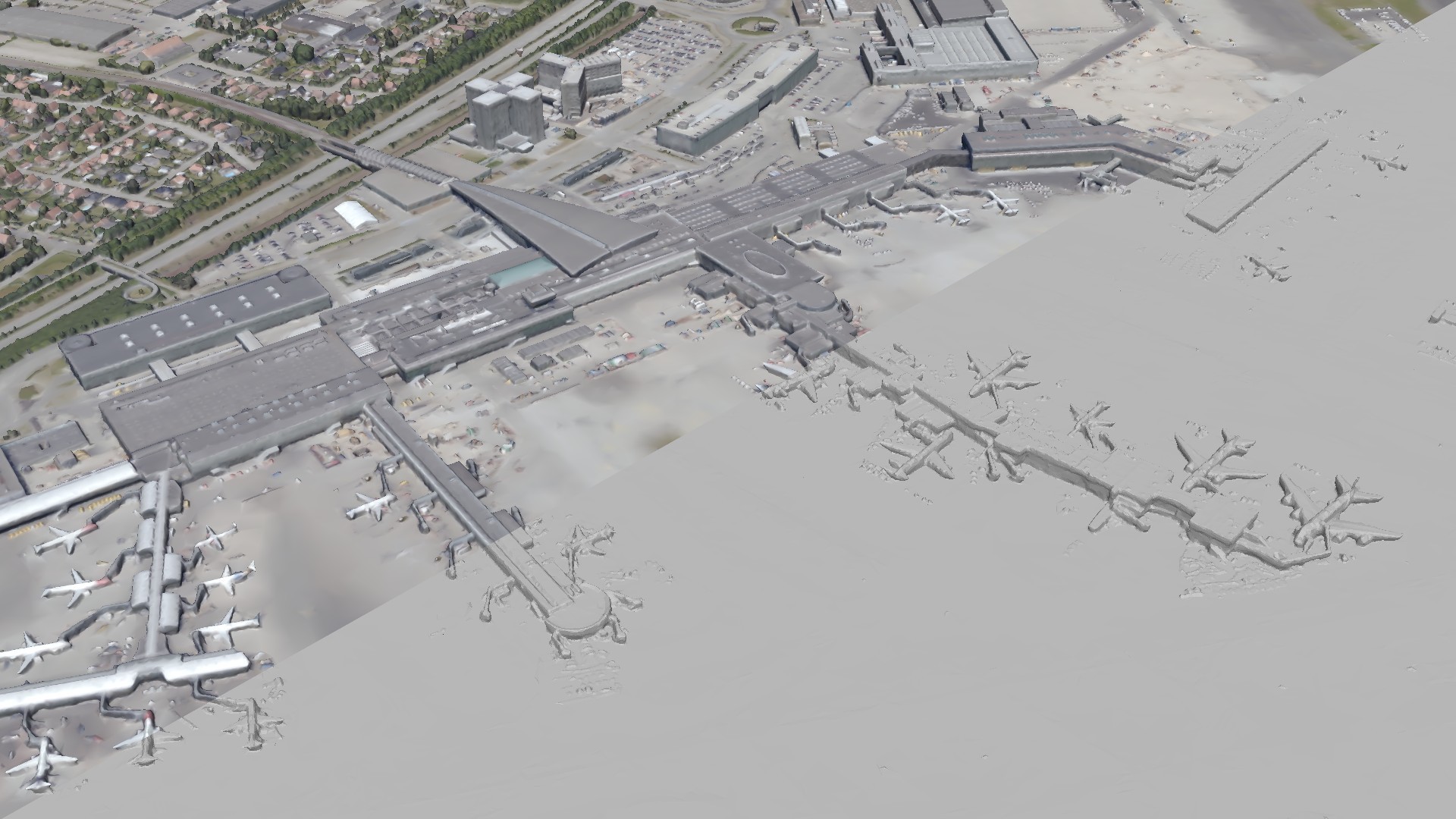}
    \end{minipage}
    \caption{Airport closeup.}
    \label{fig:closeup_airport}
\end{figure}

\begin{figure}
    \centering
    \capstart
    \begin{minipage}[b]{1.0\linewidth}
        \includegraphics[width=\textwidth]{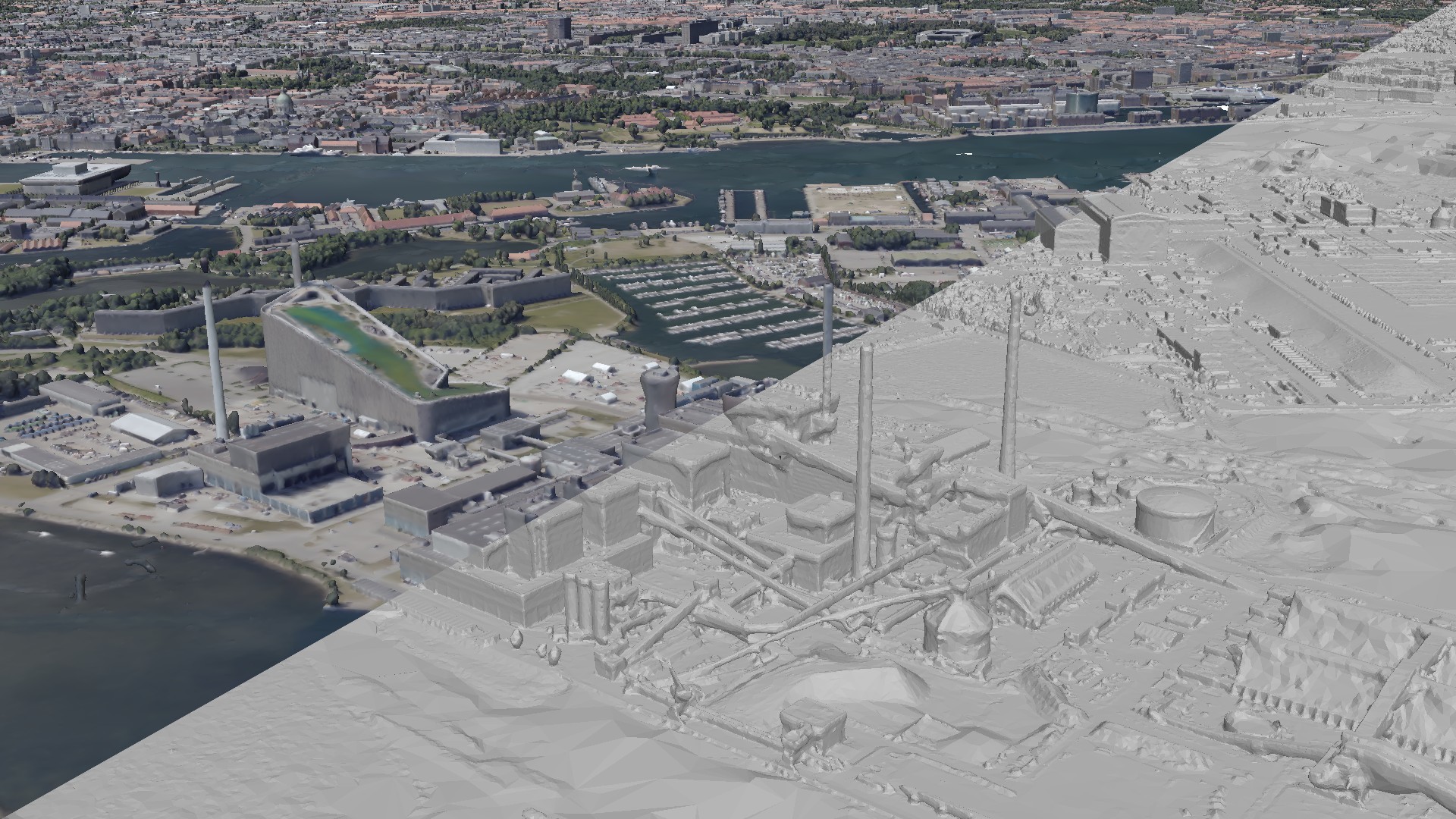}
    \end{minipage}
    \caption{Hofor Amagerverket closeup.}
    \label{fig:closeup_hofor}
\end{figure}

\begin{figure}
    \centering
    \capstart
    \begin{minipage}[b]{1.0\linewidth}
        \includegraphics[width=\textwidth]{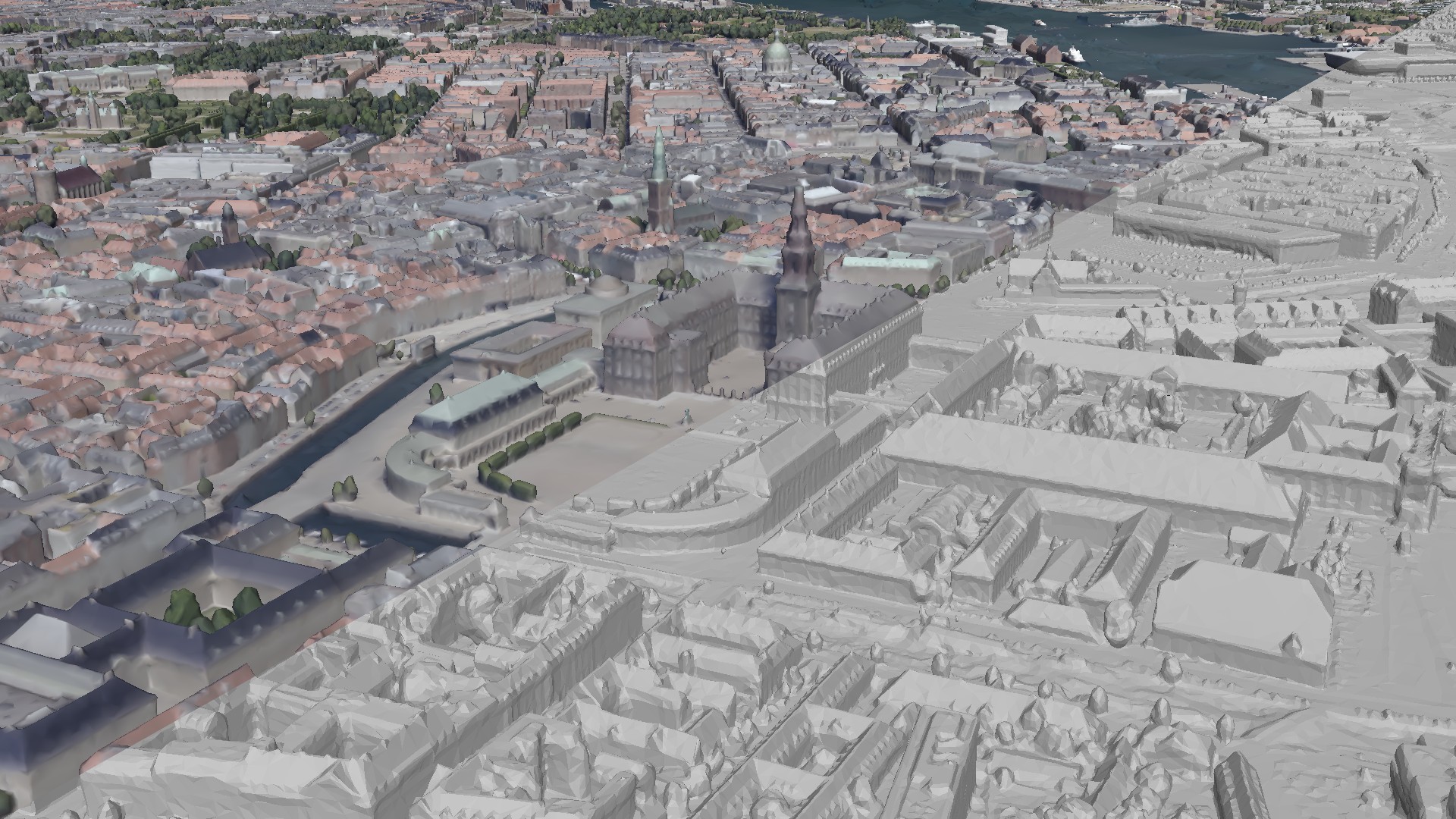}
    \end{minipage}
    \caption{Christiansborg Palace closeup.}
    \label{fig:closeup_christiansborg_palace}
\end{figure}

\begin{figure}
    \centering
    \capstart
    \begin{minipage}[b]{1.0\linewidth}
        \includegraphics[width=\textwidth]{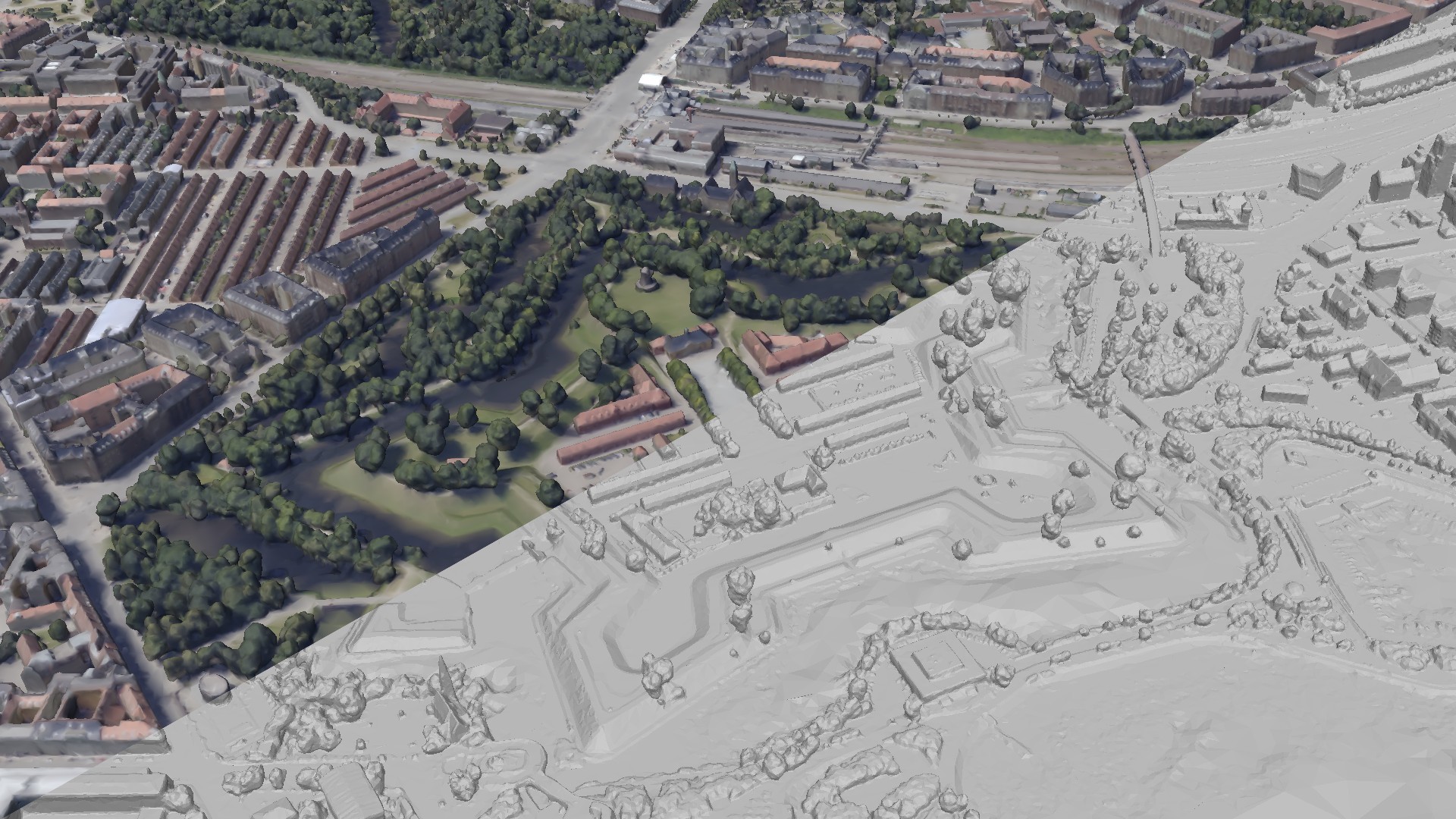}
    \end{minipage}
    \caption{Kastellet closeup.}
    \label{fig:closeup_kastellet}
\end{figure}

\begin{figure}
    \centering
    \capstart
    \begin{minipage}[b]{1.0\linewidth}
        \includegraphics[width=\textwidth]{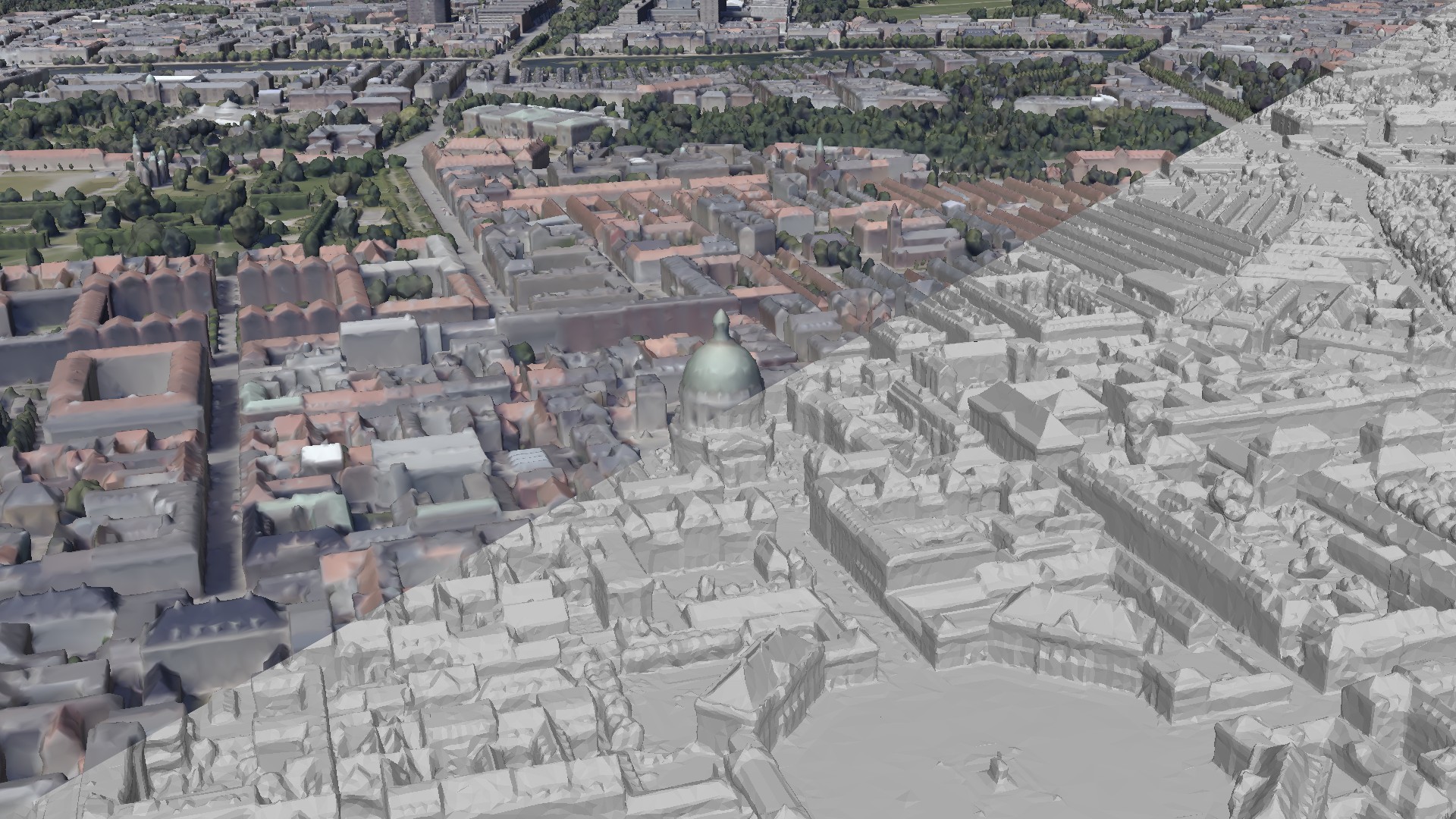}
    \end{minipage}
    \caption{Frederik's Church closeup.}
    \label{fig:closeup_frederik_church}
\end{figure}

\begin{figure}
    \centering
    \capstart
    \begin{minipage}[b]{1.0\linewidth}
        \includegraphics[width=\textwidth]{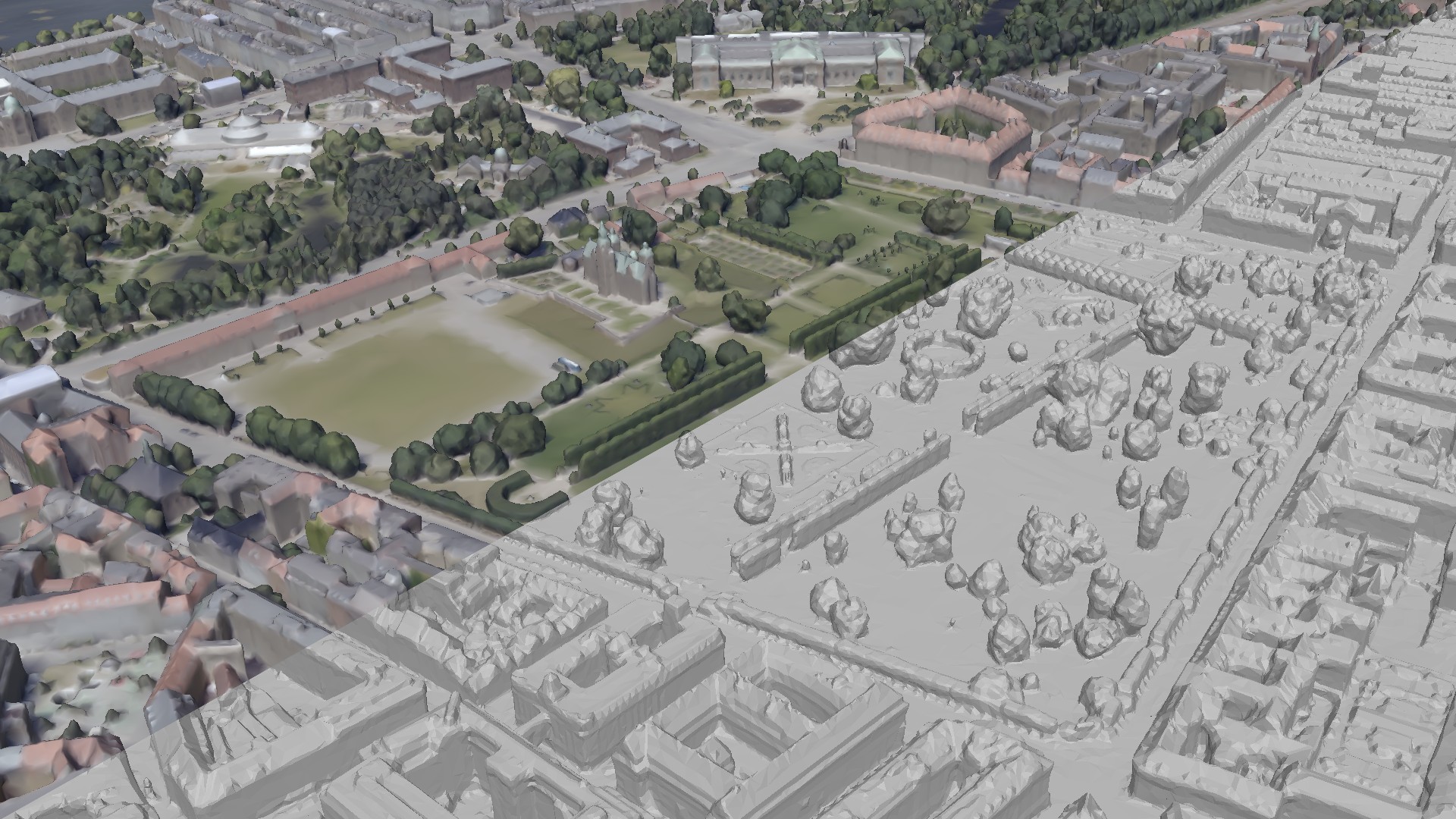}
    \end{minipage}
    \caption{Rosenborg Castle closeup.}
    \label{fig:closeup_rosenborg_castle}
\end{figure}

\begin{figure}
    \centering
    \capstart
    \begin{minipage}[b]{1.0\linewidth}
        \includegraphics[width=\textwidth]{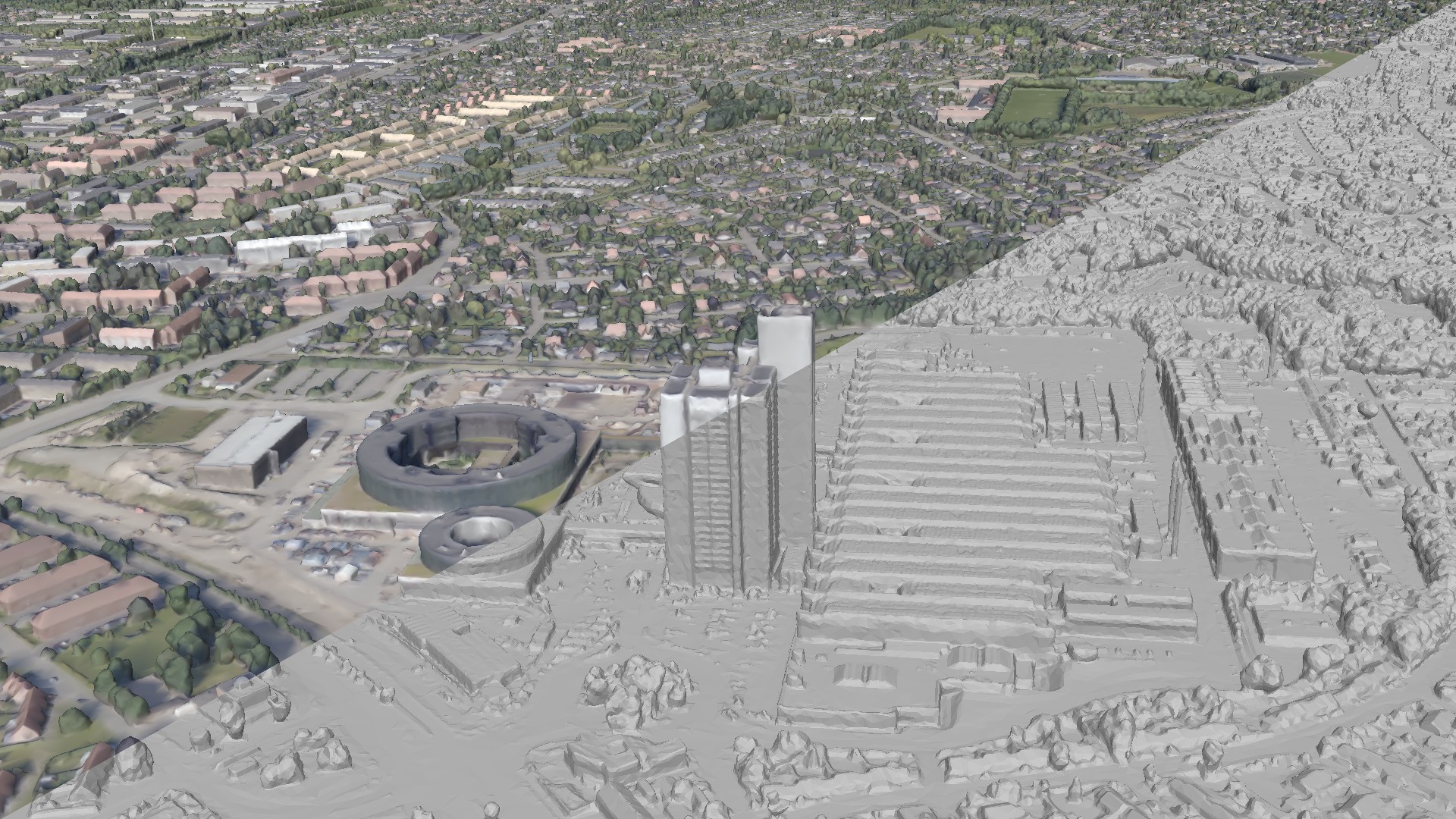}
    \end{minipage}
    \caption{Herlev Hospital closeup.}
    \label{fig:closeup_hospital}
\end{figure}

\begin{figure}
    \centering
    \capstart
    \begin{minipage}[b]{1.0\linewidth}
        \includegraphics[width=\textwidth]{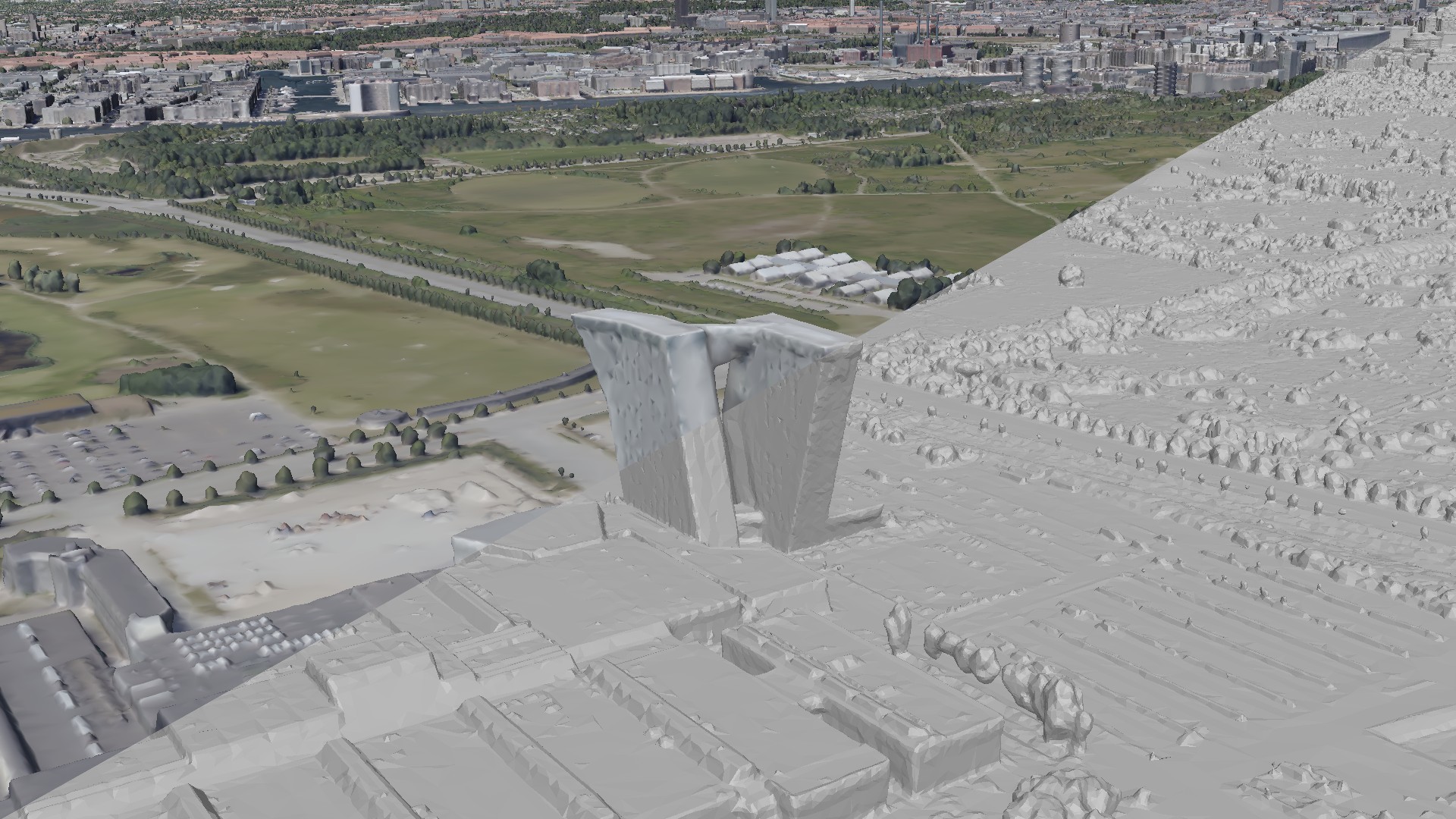}
    \end{minipage}
    \caption{AC Hotel by Marriott Bella Sky closeup.}
    \label{fig:closeup_hotel}
\end{figure}

\begin{figure}
    \centering
    \capstart
    \begin{minipage}[b]{1.0\linewidth}
        \includegraphics[width=\textwidth]{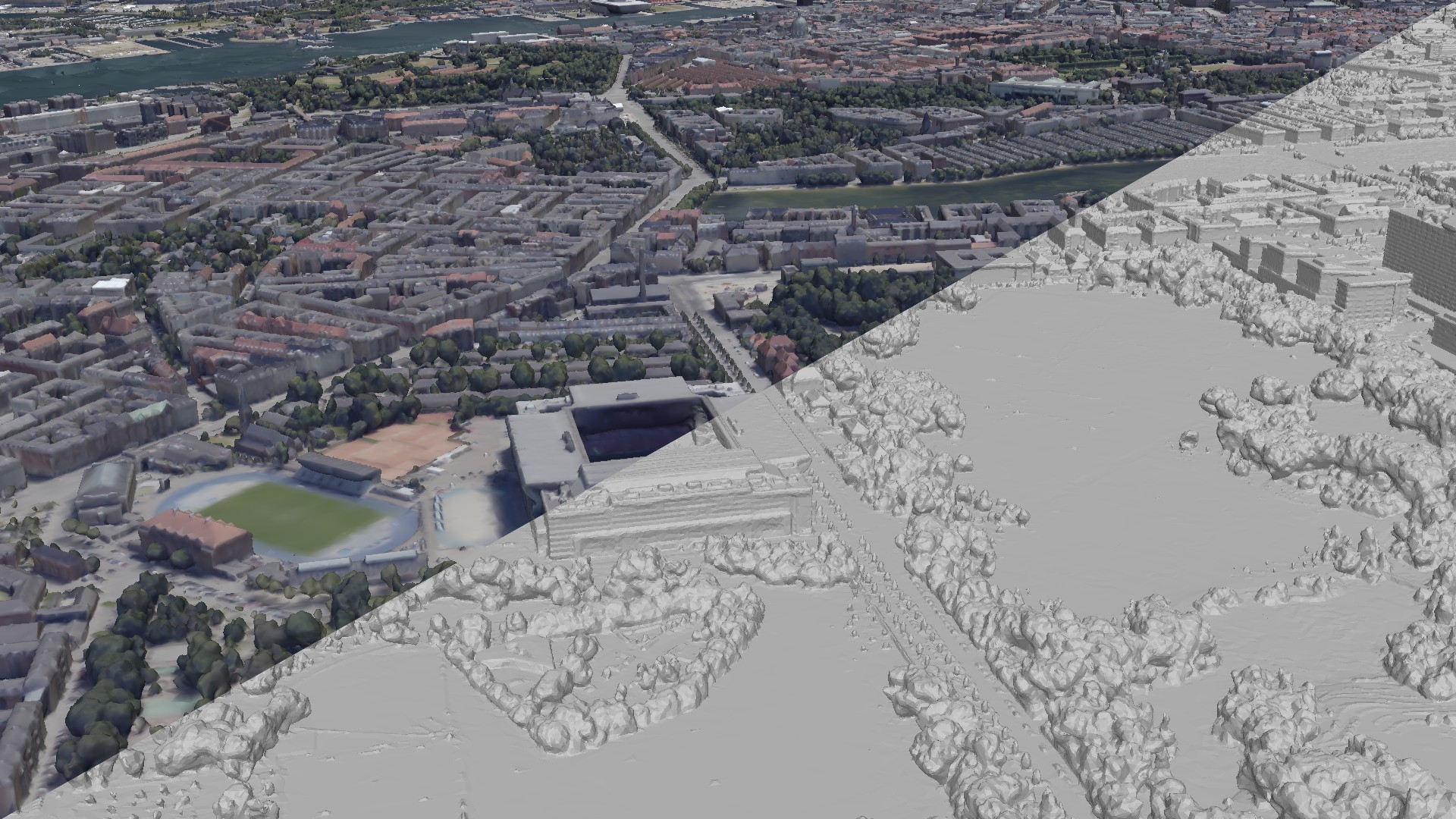}
    \end{minipage}
    \caption{Telia Parken Stadium closeup.}
    \label{fig:closeup_stadium}
\end{figure}

\begin{figure}
    \centering
    \capstart
    \begin{minipage}[b]{1.0\linewidth}
        \includegraphics[width=\textwidth]{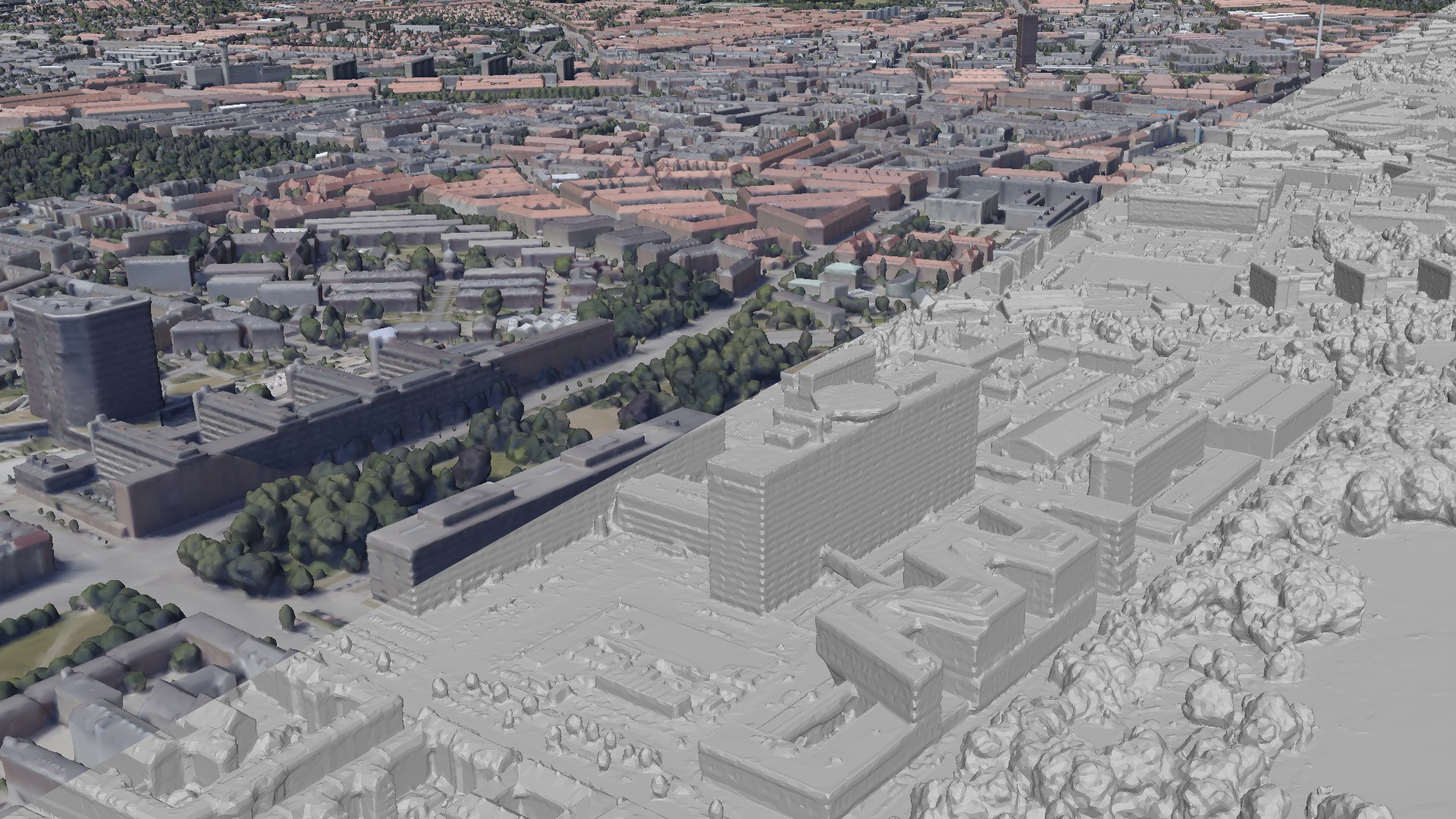}
    \end{minipage}
    \caption{Rigshospitalet closeup.}
    \label{fig:closeup_rigshospitalet}
\end{figure}

\begin{figure}
    \centering
    \capstart
    \begin{minipage}[b]{1.0\linewidth}
        \includegraphics[width=\textwidth]{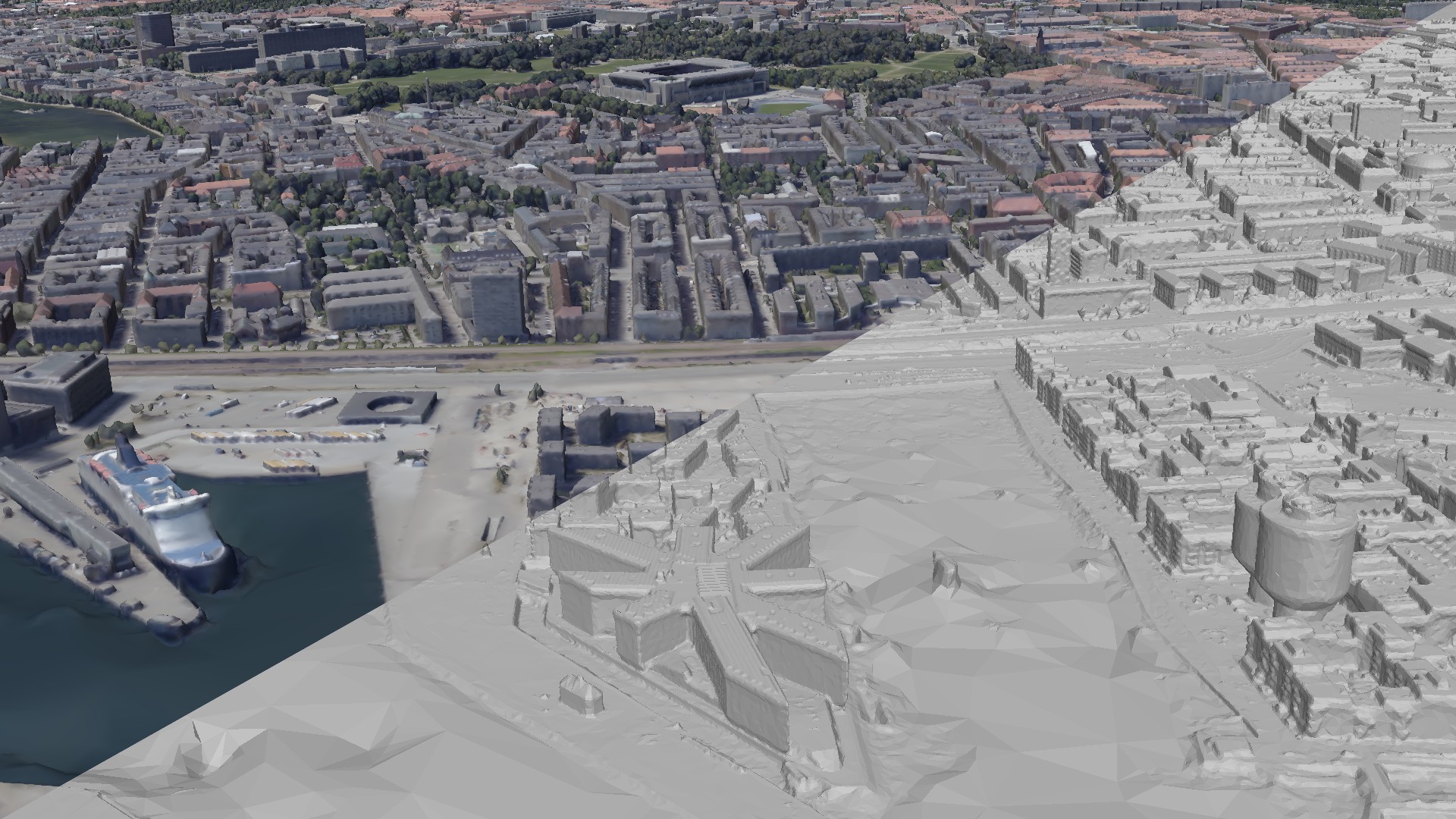}
    \end{minipage}
    \caption{UN City closeup.}
    \label{fig:closeup_un_city}
\end{figure}

\begin{figure}
    \centering
    \capstart
    \begin{minipage}[b]{1.0\linewidth}
        \includegraphics[width=\textwidth]{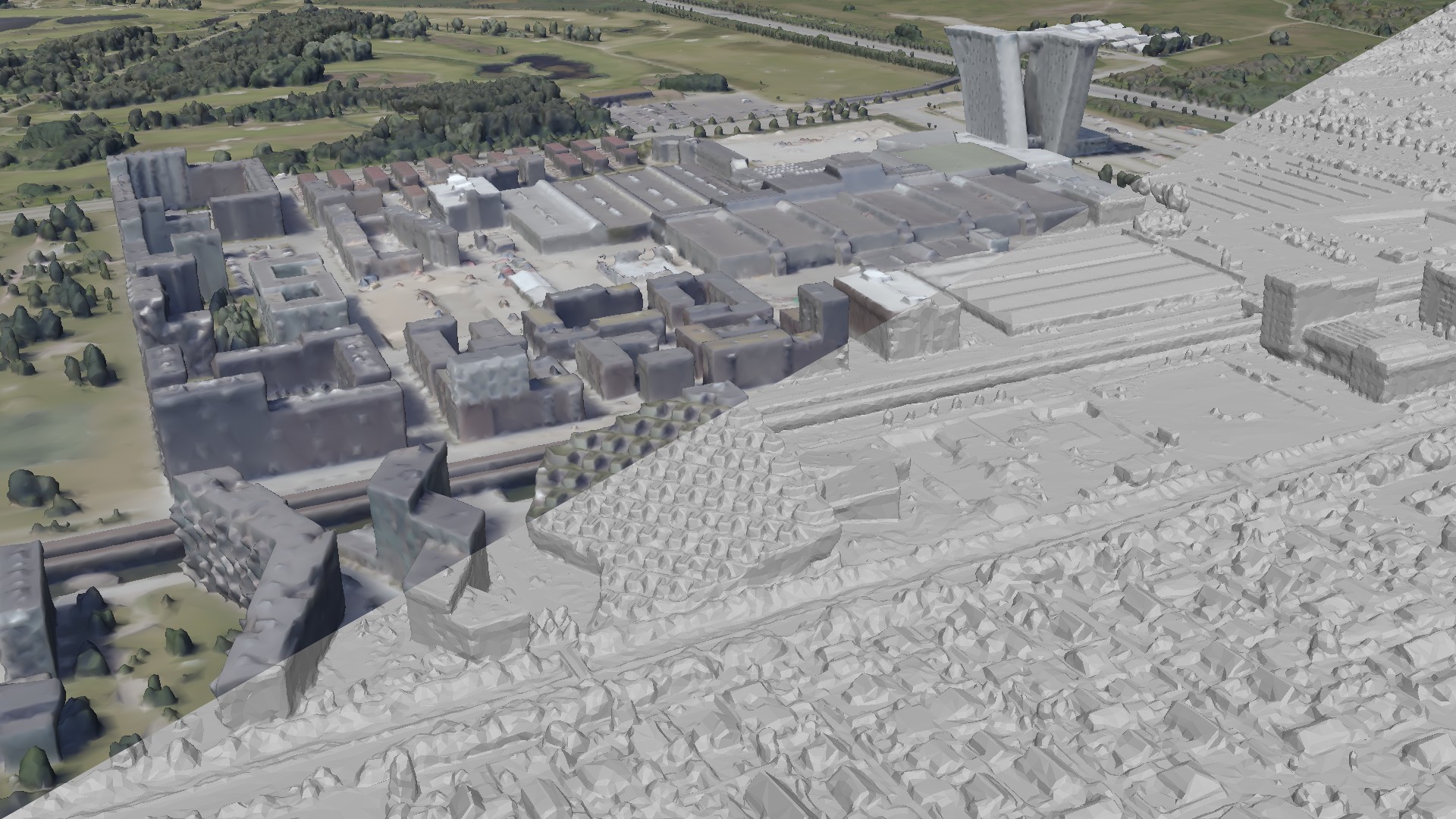}
    \end{minipage}
    \caption{Mountain Dwellings closeup.}
    \label{fig:closeup_mountain_dwellings}
\end{figure}

\clearpage
{\small
\bibliographystyle{ieee_fullname}
\bibliography{egbib}
}